\pdfoutput=1

\documentclass[11pt]{article}

\usepackage[]{acl}

\usepackage{times}
\usepackage{latexsym}

\usepackage[T1]{fontenc}

\usepackage[utf8]{inputenc}

\usepackage{microtype}

\usepackage{inconsolata}

\usepackage{amsmath} 
\usepackage{amssymb} 
\usepackage{float}   
\usepackage{graphicx} 
\usepackage[font=small]{caption} 
\usepackage{tabularx} 
\usepackage{array} 
\usepackage{longtable} 
\usepackage{hyperref} 
\usepackage{siunitx} 
\usepackage{url} 
\usepackage{cuted} 
\usepackage{enumitem}
\usepackage{wrapfig}
\usepackage{amsfonts}
\usepackage{algorithm}
\usepackage{algorithmic}
\usepackage{subcaption}
\usepackage{booktabs}
\usepackage{multirow}
\usepackage{multicol}
\usepackage{pifont}  
\usepackage{colortbl}
\usepackage[capitalise,nameinlink,noabbrev]{cleveref}
\usepackage{xspace}

\usepackage{listings}
\usepackage{xcolor}
\usepackage{setspace}
\usepackage[most]{tcolorbox}
\usepackage{wrapfig}    
\definecolor{inputcolor}{RGB}{26, 115, 232}   
\definecolor{outputcolor}{RGB}{233, 89, 33}   
\definecolor{gtcolor}{RGB}{0, 102, 51}        
\definecolor{boxbg}{RGB}{245, 245, 245}       

\newcommand{\dataset}{LMOD\xspace}
\newcommand{\datasetlong}{Large Multimodal Ophthalmology Dataset\xspace}

\title{
\begin{minipage}{.05\textwidth}
\centering
\vspace{-4pt}
\includegraphics[width=\linewidth]{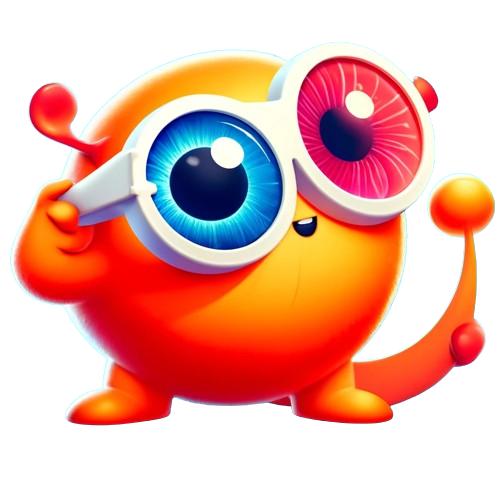} 
\end{minipage}
\dataset: A \datasetlong and \\ Benchmark for Large Vision-Language Models
}

\author{
\vspace{2mm} 
        Zhenyue Qin$^1$\thanks{~~Equal contribution}\hspace{3mm}
        Yu Yin$^2$\footnotemark[1] \hspace{3mm}
        Dylan Campbell$^3$ \hspace{3mm}
        Xuansheng Wu$^4$ 
    \\
    \textbf{
        Ke Zou$^5$ \hspace{3mm}
        Yih-Chung Tham$^5$ \hspace{3mm}
        Ninghao Liu$^4$ \hspace{3mm}
        Xiuzhen Zhang$^6$ \hspace{3mm}
        Qingyu Chen$^1\thanks{~~Correspondance email: qingyu.chen@yale.edu}$
    }
    \vspace{2mm}
    \\
    \hspace{-3mm}
    \textsuperscript{1}Yale University  
    \hspace{4mm}  
    \textsuperscript{2}Imperial College London  
    \hspace{4mm}  
    \textsuperscript{3}Australian National University
    \\
    \textsuperscript{4}University of Georgia \hspace{4mm}
    \textsuperscript{5}National University of Singapore \hspace{4mm}
    \textsuperscript{6}RMIT University
    \vspace{2mm}
    \\
    \hspace{-8mm}
    \centering
    Project Page: \url{https://kfzyqin.github.io/lmod/}
}

\begin{document}
\maketitle

\begin{abstract}
The prevalence of vision-threatening eye diseases is a significant global burden, with many cases remaining undiagnosed or diagnosed too late for effective treatment. Large vision-language models (LVLMs) have the potential to assist in understanding anatomical information, diagnosing eye diseases, and drafting interpretations and follow-up plans, thereby reducing the burden on clinicians and improving access to eye care. However, limited benchmarks are available to assess LVLMs' performance in ophthalmology-specific applications. In this study, we introduce \dataset, a large-scale multimodal ophthalmology benchmark consisting of 21,993 instances across (1) five ophthalmic imaging modalities: optical coherence tomography, color fundus photographs, scanning laser ophthalmoscopy, lens photographs, and surgical scenes; (2) free-text, demographic, and disease biomarker information; and (3) primary ophthalmology-specific applications such as anatomical information understanding, disease diagnosis, and subgroup analysis. In addition, we benchmarked 13 state-of-the-art LVLM representatives from closed-source, open-source, and medical domains. The results demonstrate a significant performance drop for LVLMs in ophthalmology compared to other domains. Systematic error analysis further identified six major failure modes: misclassification, failure to abstain, inconsistent reasoning, hallucination, assertions without justification, and lack of domain-specific knowledge. In contrast, supervised neural networks specifically trained on these tasks as baselines demonstrated high accuracy. These findings underscore the pressing need for benchmarks in the development and validation of ophthalmology-specific LVLMs.

\end{abstract}

\section{Introduction}

\begin{figure}[tp]
    \centering
    \includegraphics[width=1\linewidth]{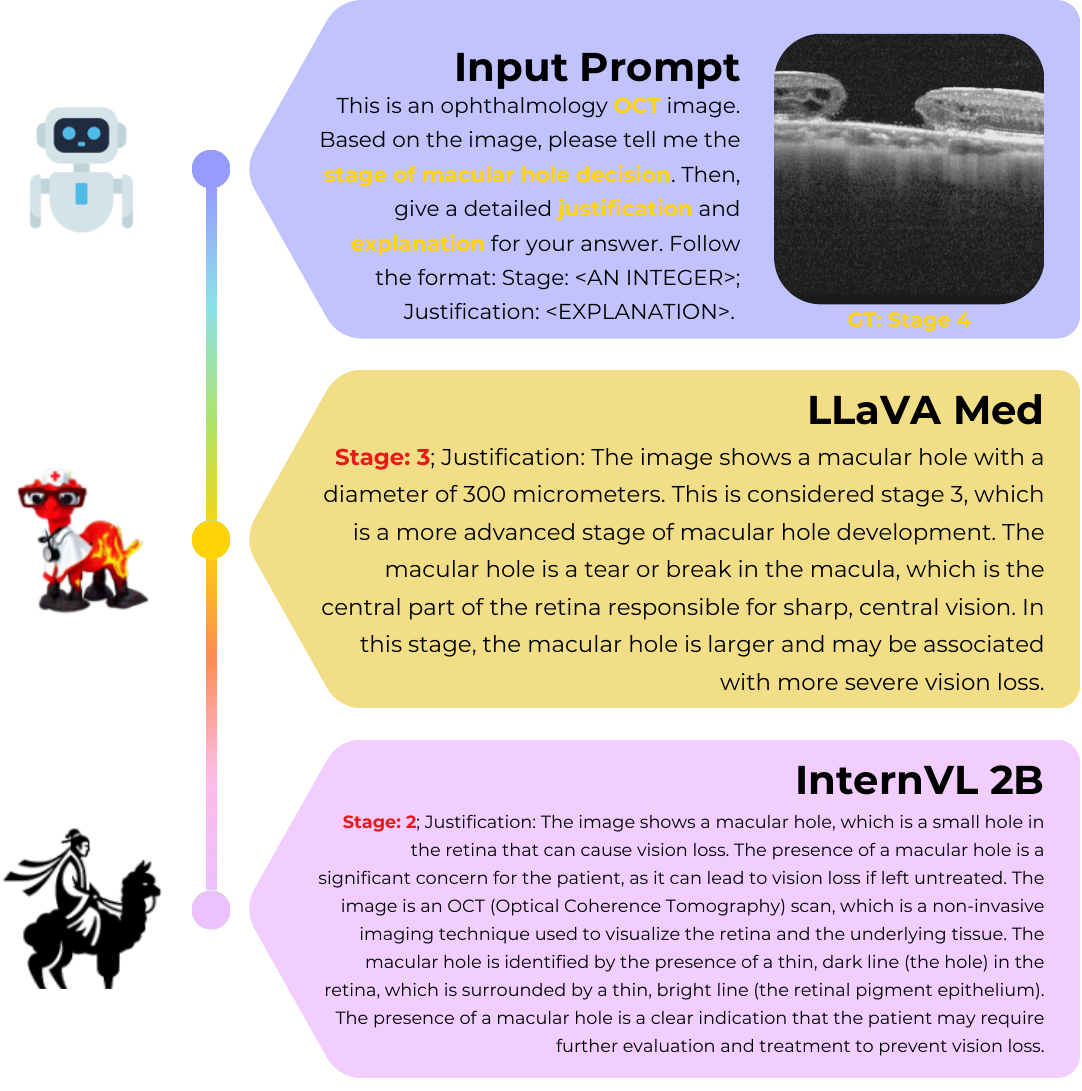}
    \caption{LVLM response examples for macular hole staging.}
    \vspace{-5pt}
    \label{fig:example_responses}
\end{figure}

\begin{figure*}[tp]
    \centering
    \includegraphics[width=1\linewidth]{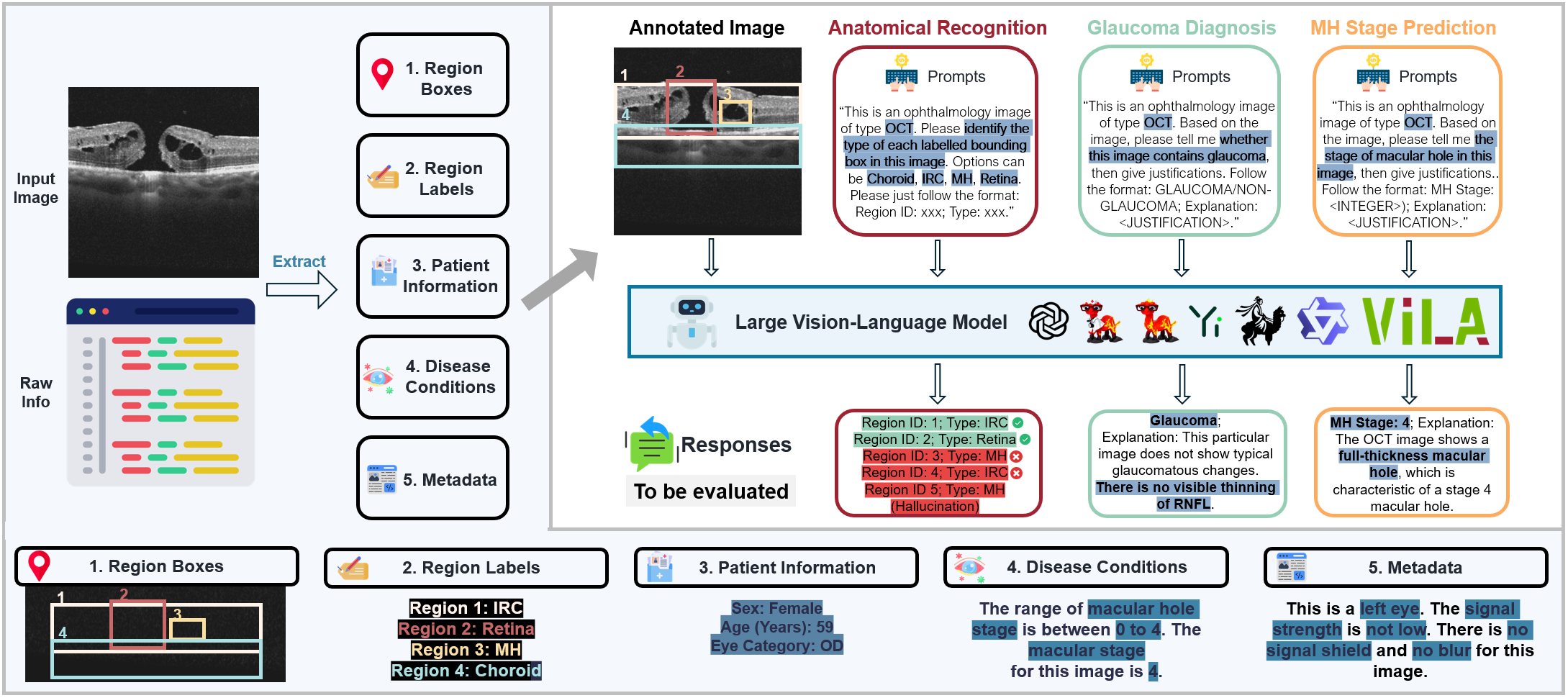}
    \caption{Overview of our data processing and evaluation pipeline for assessing the performance of LVLMs. The raw information is preprocessed to extract structured data such as bounding boxes and disease conditions. This aggregated information is then used to generate prompts for the LVLMs to identify the type of each labeled region, or conduct diagnosis analysis. The LVLMs processes the input image and prompt to generate responses categorizing each region or disease or describing diseases. Finally, the model's output is compared against the ground truth results using our proposed evaluation metrics. }
    \label{fig:overall_pipeline}
\vspace{-5pt}
\end{figure*}

The ever-increasing prevalence of primary eye diseases poses a significant global burden, with more than 2.2 billion individuals suffering from vision impairment worldwide, particularly in low- and middle-income regions \cite{tham2014global,neely2017prevalence,cavan2017diabetic,2023_eye_who}. Limited access to comprehensive eye examinations and a shortage of clinicians result in a substantial proportion of major eye diseases being undiagnosed or diagnosed too late for effective treatment \cite{neely2017prevalence,varma2011assessment}. To address this challenge, artificial intelligence (AI) applications, particularly Large Language Models (LLMs) and their multimodal extensions, have been developed to streamline clinical workflows by assisting in patient triage, disease diagnosis, prognosis prediction, and reducing documentation burdens \cite{ting2019artificial,peng2019deepseenet,keenan2022deeplensnet,2018_nm_clinically,2023_nature_eye_foundation,2023_eye_federated,2024_eye_classification,tian2024opportunities,chen2023large,qin2022medical}. These large vision-language models (LVLMs), such as GPT-4o \cite{2023_openai_gpt4} and LLaVA \cite{2024_neurips_llava}, combine LLMs with vision encoders to generate responses based on input images, which are particularly beneficial in medical imaging, where interpretations and follow-up plans are documented in free-text clinical notes \cite{tian2024opportunities,xiao2024comprehensive}.

Specifically, in ophthalmology, LVLMs enable (1) automated disease diagnosis and classification, such as assessing the severity level of glaucoma; (2) region classification and segmentation, such as segmenting retinal nerve fiber layers in optical coherence tomography (OCT); and (3) documentation, such as generating free-text imaging interpretations \cite{betzler2023large}. Importantly, what distinguishes LVLMs is their ability to handle these tasks within a single model using zero- or few-shot learning. In contrast, previous applications required sophisticated fine-tuning on substantial manually curated instances, making it challenging to adapt fine-tuned models to different data types \cite{xiao2024comprehensive,betzler2023large}. This may improve the efficiency and accuracy of workflows in ophthalmology clinics \cite{2018_nm_clinically}.

\begin{table*}[!t]
\centering
\setlength\tabcolsep{6pt}
\resizebox{1\linewidth}{!}{
\begin{tabular}{@{}lccccccccccc@{}}
\toprule
\multirow{3}{*}{\textbf{Benchmarks}} & \multicolumn{2}{c}{\textbf{Modalities}} & \multicolumn{5}{c}{\textbf{Image Types}} & \multicolumn{2}{c}{\textbf{Evaluation Perspectives}} \\
\cmidrule(lr){2-3} \cmidrule(lr){4-8} \cmidrule(l){9-10} 
& \multirow{2}{*}{Images} & \multirow{2}{*}{Texts} & Surgical & \multirow{2}{*}{SLO} & \multirow{2}{*}{OCT} & Eye & Fundus & Anatomical & Diagnosis \\
& & & Scenes & & & Photos & Images & Understanding & Analysis \\
\midrule
\multicolumn{10}{c}{\textbf{General-Domain Benchmarks}} \\
\midrule
MMMU~\cite{2024_mmmu_cvpr} & $\checkmark$ & $\checkmark$ & \ding{55} & \ding{55} & \ding{55} & $\checkmark$ & $\checkmark$ & \ding{55} & \ding{55} \\
MME-RealWorld~\cite{2024_mme} & $\checkmark$ & $\checkmark$ & \ding{55} & \ding{55} & \ding{55} & \ding{55} & \ding{55} & \ding{55} & \ding{55} \\
UNK-VQA~\cite{2024_unkvqa_tpami} & $\checkmark$ & $\checkmark$ & \ding{55} & \ding{55} & \ding{55} & \ding{55} & \ding{55} & \ding{55} & \ding{55} \\
MMCBench~\cite{2024_benchlmmcor} & $\checkmark$ & $\checkmark$ & \ding{55} & \ding{55} & \ding{55} & \ding{55} & \ding{55} & \ding{55} & \ding{55} \\
MathVista~\cite{2023_mathvista_iclr} & $\checkmark$ & $\checkmark$ & \ding{55} & \ding{55} & \ding{55} & \ding{55} & \ding{55} & \ding{55} & \ding{55} \\
SEED-Bench~\cite{2024_seed_cvpr} & $\checkmark$ & $\checkmark$ & \ding{55} & \ding{55} & \ding{55} & \ding{55} & \ding{55} & \ding{55} & \ding{55} \\
\midrule
\multicolumn{10}{c}{\textbf{Ophthalmology-Specific Benchmarks}} \\
\midrule
Bench-Nephrology~\cite{2024_benchnephrology_nejmai} & \ding{55} & $\checkmark$ & \ding{55} & \ding{55} & \ding{55} & \ding{55} & \ding{55} & \ding{55} & \ding{55} \\
Eval-GPT-Ophth~\cite{2023_evalgptophth_ophsci} & \ding{55} & $\checkmark$ & \ding{55} & \ding{55} & \ding{55} & \ding{55} & \ding{55} & \ding{55} & \ding{55} \\
Bench-Myopia~\cite{2023_benchmyopia_ebio} & \ding{55} & $\checkmark$ & \ding{55} & \ding{55} & \ding{55} & \ding{55} & \ding{55} & \ding{55} & \ding{55} \\
OphNet~\cite{2024_ophnet} & $\checkmark$ & $\checkmark$ & $\checkmark$ & \ding{55} & \ding{55} & \ding{55} & \ding{55} & \ding{55} & \ding{55} \\
\midrule 
\textbf{\dataset (ours)} & $\checkmark$ & $\checkmark$ & $\checkmark$ & $\checkmark$ & $\checkmark$ & $\checkmark$ & $\checkmark$ & $\checkmark$ & $\checkmark$ \\
\bottomrule
\end{tabular}%
}
\caption{Comparison of existing general-domain and ophthalmology-specific benchmarks for evaluating large vision-language models, highlighting their supported modalities, coverage of image types, and evaluation perspectives.}
\label{tab:benchmark_comparison}
\end{table*}

Nevertheless, limited benchmarks are available to assess LVLMs' performance in ophthalmology-specific applications to date, and consequently, the potential and limitations of LVLMs are not clear. Existing studies in ophthalmology focus on evaluating the performance of LLMs on text-based tasks, including multiple-choice questions \cite{2024_benchnephrology_nejmai}, general ophthalmology knowledge testing \cite{2023_evalgptophth_ophsci}, and free-text question answering on specific ophthalmology topics \cite{2023_benchmyopia_ebio}. While these studies are useful for demonstrating potential, ophthalmic images are arguably the most important data modality. Ophthalmologists require different imaging modalities for diagnosis and prognosis that are not derived from the text itself \cite{khan2021global}. Therefore, the capability of image analysis is crucial in the domain of ophthalmology. However, this capability poses challenges for existing LVLMs, which struggle to understand and analyze ophthalmic images, as illustrated in \cref{fig:example_responses}. In addition, existing ophthalmic imaging datasets were designed for the development and evaluation of fine-tuning AI models; they may only contain a single imaging modality (e.g., OCT only), a specific task (e.g., region segmentation), and specific output types (e.g., a disease severity class rather than free text). 

In response, this study proposes a systematic and reproducible data and evaluation pipeline that repurposes existing datasets to curate a dataset we refer to as \dataset (Large Multimodal Ophthalmology Dataset) for the development and evaluation of LVLMs in ophthalmology. \dataset consists of five imaging modalities: surgical scenes (SS), optical coherence tomography (OCT), scanning laser ophthalmoscopy (SLO), lens photographs (LP), and color fundus photographs (CFP), collectively comprising over 20K instances. It also provides multi-granular annotations, including region annotation and disease information. 
The pipeline can also be directly applied to new datasets. 



Our contributions are as follows: (1) We introduce \dataset, a large-scale ophthalmology dataset that includes over 21K images across diverse imaging modalities. \dataset is richly annotated with disease labels and bounding boxes, supporting comprehensive evaluations from macro-level. (2) We systematically benchmark 13 state-of-the-art (SoTA) LVLMs, including models with diverse visual backbones and LLMs. The evaluation is conducted using a wide range of metrics, assessing strengths and weaknesses of LVLMs from various perspectives. (3) Through fine-tuning and supervised classification, we demonstrate that while the challenges posed by ophthalmic image analysis are intricate for LVLMs, they are insurmountable. Our comprehensive evaluations and error analysis provide both a high-level overview and detailed insights, presented through various result formats, including weighted averages, bar charts, radar charts, and visual illustrations, to highlight the key strengths and weaknesses.



\section{Related Work}

This section provides an overview of the advancements in LVLMs and highlights the lack of comprehensive benchmarks in ophthalmology. 

\subsection{Advances in LVLMs}

The release of ChatGPT~\cite{2023_openai_gpt4} has sparked considerable interest in the potential of large language models (LLMs) across various domains~\cite{liu2023summary, tian2024opportunities, de2023chatgpt}. Building on the success of ChatGPT and other LLMs~\cite{radford2018improving,radford2019language,brown2020language}, researchers have developed LVLMs that integrate the strengths of vision encoders with LLMs. These models employ vision encoders, typically pretrained on vast image datasets in an unsupervised manner, to extract visual features from images and incorporate them into LLMs, enabling a combined understanding of both vision and language. Several notable LVLMs have been introduced in recent years, each with its unique architecture or training approach, such as GPT-4~\cite{2023_openai_gpt4}, LLaVA~\cite{2024_neurips_llava}, InternVL~\cite{2024_internvl_cvpr}, Qwen~\cite{bai2023qwen}, and VILA~\cite{2023_vila}. In the medical domain, representative LVLMs include LLaVA-Med \cite{2024_neurips_llavamed} and its variants~\cite{2024_neurips_llavamed, jiang2024moe, xie2024medtrinity}, which demonstrate potential for disease diagnosis and medical question answering. The advent of LVLMs has opened up new possibilities for multimodal reasoning and comprehension, with applications spanning various fields, including the medical domain~\cite{2023_llmmedicine_medcom,2023_llmusmle_plosdighealth,2023_gptmedhigheredu_radiography}.

\subsection{Lack of Benchmarks}

In ophthalmology, domain-specific foundation models, such as vision encoders pretrained on ophthalmic images, have shown consistent improvements in diagnosing and predicting the prognosis of eye diseases~\cite{2023_nature_eye_foundation,2023_eye_federated,2024_eye_classification}. However, these encoders lack the reasoning and conversational capabilities inherent to large language models (LLMs) and require task-specific fine-tuning with static inputs and outputs. Furthermore, existing research on LLMs in ophthalmology primarily focuses on text-based applications~\cite{2024_benchnephrology_nejmai,2023_evalgptophth_ophsci,2023_benchmyopia_ebio, gilson2024language}, neglecting ophthalmic images as a key data modality. To the best of our knowledge, few benchmarks exist for the development and evaluation of LVLMs in ophthalmology, posing a significant barrier to systematically evaluating the feasibility of applying existing LVLMs in this domain and hindering the development of ophthalmology-specific LVLMs. In contrast, several benchmarks have been established in both general and medical domains (\cref{tab:benchmark_comparison}). However, existing benchmarks in ophthalmology are primarily designed for the development and evaluation of AI models under the fine-tuning paradigm, often focusing on a single imaging modality, a specific task, and restricted output types. Few benchmarks encompass diverse ophthalmic imaging modalities or support a broad range of downstream evaluations, such as anatomical understanding~\cite{2021_nc_annotation} and diagnostic analysis.
\begin{table}[tp]
\vspace{3pt}
\centering
\scriptsize
\setlength{\tabcolsep}{4pt}
\resizebox{0.48\textwidth}{!}{
\begin{tabular}{@{}lccccrr@{}}
\toprule
\multirow{2}{*}{\textbf{Models}} & \multicolumn{4}{c}{\textbf{Anatomical Recognition}} & \multicolumn{2}{c}{\textbf{Diagnosis Analysis}} \\ \cmidrule(lr){2-5} \cmidrule(lr){6-7} 
 & \multirow{2}{*}{Precision $\uparrow$} & \multirow{2}{*}{Recall $\uparrow$} & \multirow{2}{*}{F1 $\uparrow$} & \multirow{2}{*}{HC $\uparrow$} & Glaucoma & MH Stage \\
 & & & & & Acc (\%) $\uparrow$ & Acc (\%) $\uparrow$ \\ \midrule
Random & - & - & - & - & 50.00 & 25.00 \\
\midrule
Finetuned & Invalid & Invalid & Invalid & Invalid & Invalid & Invalid \\
\midrule 
GPT-4o & 0.5609 & \textbf{0.5896} & \textbf{0.5748} & 0.9513 & \textbf{54.09} & 19.71 \\
LLaVA-Med & 0.0789 & 0.1163 & 0.0940 & 0.7435 & 50.00 & 25.00 \\
LLaVA-1.5-7B & 0.0567 & 0.0410 & 0.0475 & 0.2674 & 50.00 & 7.30 \\
LLaVA-M-7B & 0.1346 & 0.1450 & 0.1396 & 0.7569 & 50.00 & 0.00 \\
LLaVA-V-7B & 0.3095 & 0.2540 & 0.2790 & 0.7516 & 50.00 & 0.00 \\
LLaVA-13B & 0.0599 & 0.0803 & 0.0686 & 0.5993 & 50.00 & 0.00 \\
Yi-6B & 0.1952 & 0.1499 & 0.1695 & 0.8893 & 50.00 & 5.26 \\
InternVL-2B & 0.6025 & 0.3999 & 0.4807 & \textbf{0.9809} & 50.00 & \textbf{30.26} \\
InternVL-4B & \textbf{0.7241} & 0.4481 & 0.5536 & 0.9629 & 50.00 & 25.00 \\
Qwen & 0.0275 & 0.0372 & 0.0316 & 0.8418 & 50.00 & 18.42 \\
VILA-3B & 0.1429 & 0.1119 & 0.0633 & 0.5300 & 50.00 & 24.24 \\
VILA-3B-S2 & 0.3340 & 0.2636 & 0.1255 & 0.7695 & 50.00 & 21.42 \\
VILA-8B & N/A & N/A & N/A & N/A & 50.00 & 22.53 \\
\midrule
Average & 0.2688 & 0.2197 & 0.2189 & 0.7537 & 50.31 & 15.31 \\
\bottomrule
\end{tabular}
}
\caption{Performance comparison of state-of-the-art large vision-language models on the \dataset benchmark, evaluating their capabilities in anatomical recognition and diagnosis analysis. Acc indicates accuracy. The best-performing model in each metric is highlighted in bold. LLaVA-Med-Finetuned consistently produced invalid outputs across both tasks.}
\label{tab:performance_comparison}
\end{table}

\vspace{-5pt}
\section{\dataset Curation}
\vspace{-5pt}

In this section, we present our methodology for curating \dataset: \datasetlong. We first describe our data curation pipeline, which involves selecting suitable datasets, generating consistent annotations across various image types, and designing standardized prompts for model evaluation. Our ophthalmology clinicians, with expertise in visual impairment and age-related eye diseases, were directly involved in the dataset selection, focusing on clinical relevance and diversity, and in manually defining the evaluation tasks, such as glaucoma diagnosis and macular hole staging. Next, we introduce the evaluation tasks and metrics used to assess the performance of LVLMs on \dataset, focusing on anatomical recognition and diagnostic analysis. 

\vspace{-3pt}
\subsection{Data Curation Pipeline}
\vspace{-3pt}

\begin{algorithm}[!t]
\captionsetup{font=small}
\caption{\scriptsize Anatomical Recognition Pipeline}
\label{alg:data_curation}
\begin{scriptsize} 
\begin{algorithmic}[1]
\REQUIRE Original dataset \\ $D = \{(I_1, R_1), (I_2, R_2), \ldots, (I_n, R_n)\}$, where $I_i$ is an image and $R_i$ is the corresponding raw data
\REQUIRE Minimum bounding box area threshold $\tau \in \mathbb{R}^+$
\ENSURE Curated dataset \\
$D' = \{(I_1, B'_1, P_1), (I_2, B'_2, P_2), \ldots, (I_n, B'_n, P_n)\}$, where $B'_i$ is the set of curated bounding boxes and $P_i$ is the set of corresponding prompts for image $I_i$
\FOR{each image-raw data pair $(I_i, R_i) \in D$}
    \STATE $B_i \leftarrow \text{ExtractBoundingBoxes}(R_i)$, \\
    where $B_i = \{b_{i,1}, b_{i,2}, \ldots, b_{i,|B_i|}\}$ and $b_{i,j}$ is the $j$-th bounding box of image $I_i$
\ENDFOR
\STATE $B \leftarrow \bigcup_{i=1}^n B_i$
\STATE $B' \leftarrow \{b \in B \mid \text{area}(b) \geq \tau\}$
\FOR{each image-raw data pair $(I_i, R_i) \in D$}
    \STATE $B'_i \leftarrow \{b \in B' \mid b \text{ belongs to image } I_i\}$
    \STATE $P_i \leftarrow \emptyset$
    \FOR{each bounding box $b_{i,j} \in B'_i$}
        \STATE $\textit{id}_{i,j} \leftarrow \text{GenerateUniqueID}()$
        \STATE $\textit{color}_{i,j} \leftarrow \text{AssignDistinctColor}()$
        \STATE $\textit{prompt}_{i,j} \leftarrow \text{GeneratePrompt}(b_{i,j})$
        \STATE $P_i \leftarrow P_i \cup \{(\textit{id}_{i,j}, \textit{color}_{i,j}, \textit{prompt}_{i,j})\}$
    \ENDFOR
\ENDFOR
\STATE \textbf{return} $D'$
\end{algorithmic}
\end{scriptsize} 
\end{algorithm}

\begin{figure*}[h!]
    \centering
    \includegraphics[width=1\textwidth]{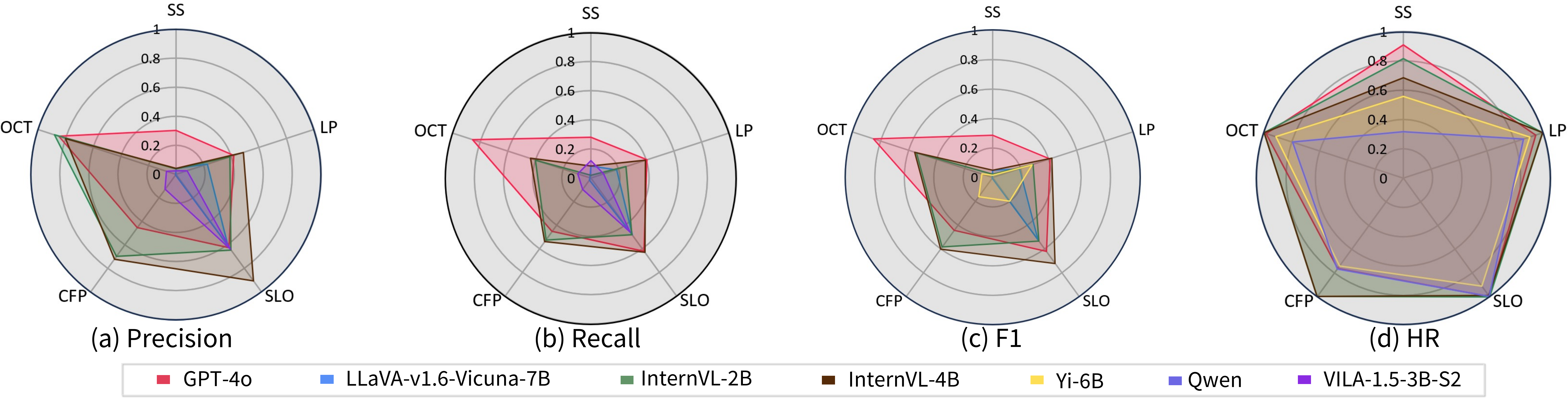}
    \caption{Performance comparison of top-performing LVLMs across different ophthalmic imaging modalities. The radar charts display the performance of the top-F1-performing models, for each evaluation metric (Precision, Recall, F1, and HR) across five different imaging modalities: surgical scenes (SS), optical coherence tomography (OCT), color fundus photographs (CFP), scanning laser ophthalmoscopy (SLO), and lens photographs (LP). }
    \label{fig:radar_chart}
\end{figure*}

The data curation pipline consists of two steps.

\textbf{Step 1: Dataset Selection.} We screened and selected the datasets for repurposing based on the following criteria: (1) \textbf{Accessibility:} The datasets are publicly accessible and non-commercial, ensuring that \dataset can be widely used by the research community for model evaluation and development. (2) \textbf{Coverage:} The datasets collectively need to cover key ophthalmic imaging modalities and primary eye diseases. (3) \textbf{Diversity:} Population diversity needs to be considered, enabling the benchmarks to quantify model effectiveness across subgroups, especially within disparity populations.

As a result, nine datasets were selected for curation. These datasets can be categorized into five ophthalmic imaging types, and the overall statistics can be found in \cref{tab:dataset_overview}:

\textbf{SS: }SS stands for surgical scenes. This category includes the Cataract-1K dataset \cite{2024_cataract1k_scidata}, which contains 2,256 images extracted from cataract surgery videos. These images capture various stages and aspects of the surgical procedure, with an average of 3.3 bounding boxes per image.

\textbf{OCT:} OCT represents Optical Coherence Tomography. OIMHS \cite{2023_oimhs_scidata} represents this category, comprising 3,859 OCT images. OCT is a non-invasive imaging technique that provides high-resolution cross-sectional retinal images. There 2.4 bounding boxes per image.

\textbf{SLO:} SLO indicates Scanning Laser Ophthalmoscopy. The Harvard FairSeg dataset \cite{2024_harvardfairseg_tmi} is included in this category, featuring 10,000 SLO fundus images. Each image contains an average of a single bounding box.

\begin{table}
\centering
\resizebox{0.5\textwidth}{!}{
\begin{tabular}{l r r}
\toprule
\textbf{Data Types} & \textbf{Num Images} & \textbf{Num Avg Boxes} \\
\midrule
Surgical Scenes (SS) & 2,256 & 3.3 \\
Optical Coherence Tomography (OCT) & 3,859 & 2.4 \\
Scanning Laser Ophthalmoscopy (SLO) & 10,000 & 1.0 \\
Eye Photos (EP) & 2,432 & 1.9 \\
Color Fundus Images (CFI) & 3,386 & 1.6 \\
\bottomrule
\end{tabular}
}
\caption{Overview of \dataset, including the number of images (Num Images) and average number of bounding boxes per image (Num Avg Boxes). }
\label{tab:dataset_overview}
\vspace{-5pt}
\end{table}

\textbf{LP:} LP means Lens Photographs. This category includes two datasets: CAU001~\cite{2023_cau001} and Cataract Detection 2~\cite{2023_cataract_det_2}. CAU001 contains 1,417 RGB photographs of human eye regions, with bounding box annotations indicating the locations of the left and right eyes, pupils, and irises. Cataract Detection 2 consists of 1,015 photographs of eyes with and without cataracts, each containing a single bounding box annotation. Each image contains 1.9 bounding boxes in average. 

\textbf{CFP:} CFP implies Color Fundus Photographs. This category includes four datasets: REFUGE \cite{2020_refuge_miccaiw}, IDRiD \cite{2018_idrid_dataport}, ORIGA \cite{2010_origa}, and G1020 \cite{2020_g1020_ijcnn}. REFUGE contains 1,200 retinal fundus images, including both glaucoma and normal eyes, with detailed annotations of optic disc and cup segmentations. IDRiD includes 516 images with pixel-level annotations of typical diabetic retinopathy lesions and normal retinal structures. ORIGA consists of 650 retinal images annotated by trained professionals, containing a comprehensive set of image features critical for glaucoma diagnosis. G1020 contains 1,020 high-resolution color fundus photographs, accompanied by detailed ground-truth annotations, including glaucoma diagnosis, optic disc and cup segmentations, and other clinically relevant measurements. On average, each image in this category contains 1.6 bounding boxes.

\textbf{Step 2: Multi-granular Annotation.}
We further curated the datasets to support the development and evaluation of LVLMs in ophthalmology applications. These applications are categorized into two main parts: (1) anatomical understanding, which involves the accurate observation and identification of ocular structures \cite{2021_nc_annotation,2018_nm_clinically}, and (2) diagnostic analysis, which requires the interpretation of visual features and patterns to assess the presence and severity of ocular diseases. The curation procedures for each application is detailed below.

\textbf{Anatomical Recognition}: Anatomical recognition refers to the ability of models to accurately identify various anatomical structures in ophthalmic images. The algorithmic pipeline is outlined in \cref{alg:data_curation}. In brief, the steps are as follows: First, we generate bounding boxes and their associated labels. The coordinates of the bounding boxes are present in the open-source datasets. To ensure consistency and standardization across the dataset, we map the region types provided in the open-source datasets to a predefined set of ophthalmology-specific region types, such as optic disk, macula, lesion, and tumor. This mapping process allows us to handle variations in terminology and granularity used in the original datasets.


To balance between the comprehensiveness of annotations and the clarity of the images, we establish a threshold and remove bounding boxes whose areas fall below this threshold (10\% for ours). This step is crucial as an excessive number of bounding boxes overlaid on the images can lead to significant occlusion, hindering the visibility and interpretability of the underlying image content. 

In the final step, using the extracted bounding box coordinates, we generate visual markers in the form of bounding boxes on the images to highlight the annotated regions. These visual markers are assigned unique labels (e.g., letters or numbers) to clearly identify each region. To increase the differentiation between regions, we assign different colors to the bounding boxes.



\begin{figure*}[t!]
    \centering
    \includegraphics[width=1\textwidth]{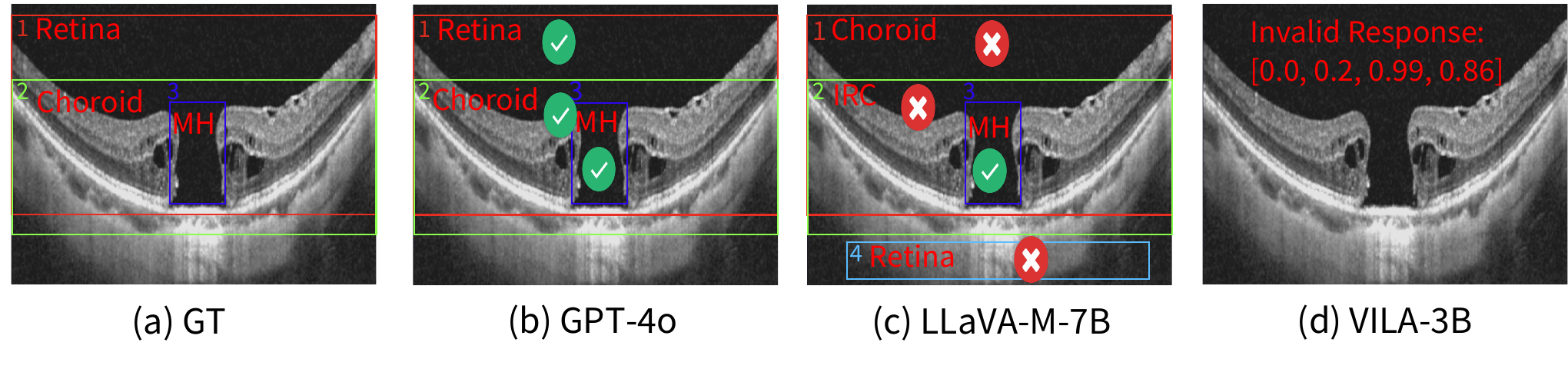}
    \caption{Visual examples of LVLM predictions for anatomical recognition in OCT images. The figure presents a comparison of ground truth (GT) annotations and predictions from three representative LVLMs: GPT-4o, LLaVA-M-7B, and VILA-3B. Green ticks indicate correct predictions, while red crosses mark incorrect ones. VILA-3B generates an invalid response consisting of a sequence of numbers unrelated to the task. }
    \label{fig:visual_example}
\vspace{-5pt}
\end{figure*}

\textbf{Diagnostic Analysis:} We evaluate models' ability to determine the presence and severity of eye diseases, using glaucoma and macular holes as case studies. We formulate glaucoma detection as a binary classification problem. Macular holes are classified into four distinct stages, ranging from 1 to 4 to indicate severity increase.







To ensure reliability and validity, we extract glaucoma and macular hole labels from the original datasets when available. In cases of imbalanced label distribution, we employ a balanced sampling strategy to mitigate potential biases and ensure fair evaluation. To be more specific, we determine the minimum number of samples across all classes and randomly select an equal number of samples from each class to create a balanced dataset. This approach guarantees that models are evaluated on a representative and unbiased sample, preventing them from exploiting class imbalance.

\begin{table}[tbp!]
    \centering
    \resizebox{0.5\textwidth}{!}{
        \begin{tabular}{lcccccccc}
        \toprule
        & \multicolumn{6}{c}{\textbf{Anatomical Recognition (F1)}} & \multicolumn{2}{c}{\textbf{Diagnosis Analysis (Accuracy)}} \\
        \cmidrule(lr){2-7} \cmidrule(lr){8-9}
        \textbf{Model} & \textbf{Macro Avg} & \textbf{SS} & \textbf{OCT} & \textbf{SLO} & \textbf{LP} & \textbf{CFP} & \textbf{Glaucoma} & \textbf{MH Stage} \\
        \midrule
        NNC & 94.36 & 93.76 & 98.42 & 94.92 & 88.85 & 95.86 & 82.69 & 98.17 \\
        \bottomrule
        \end{tabular}
    }
    \caption{\small Supervised-trained neural network classifier (NNC) performance on anatomical recognition across different image modalities and diagnosis analysis tasks. }
    \label{tab:cnn_results}
\end{table}




\vspace{-3pt}
\section{Benchmarking Results}
\vspace{-3pt}

We here present the results of benchmarking 13 state-of-the-art LVLMs on the \dataset benchmark. We conclude with an error analysis to highlight common failure modes of LVLMs.

\vspace{-2pt}
\subsection{Benchmarked LVLMs}
We benchmarked 13 LVLMs on the \dataset benchmark, including several variations of LLaVA, such as LLaVA-7B, LLaVA-M-7B (Mistral), LLaVA-V-7B (Vicuna), LLaVA-13B, and the domain-specific LLaVA-Med. Additionally, we evaluated InternVL models, including InternVL-2B and InternVL-4B, as well as VILA models, including VILA-3B, VILA-3B-S2, and VILA-8B. Other models in the benchmark included GPT-4o, Yi-6B, and Qwen. These models represent a diverse range of architectures, parameter scales, and training methodologies, highlighting the breadth of current LVLM development. See appendix for details.








\vspace{-3pt}
\subsection{Evaluation Metrics}
\vspace{-3pt}

To comprehensively assess the performance of the LVLMs on the \dataset benchmark, we employed several evaluation metrics that captured different aspects of their capabilities. These metrics provided a holistic view of the models' strengths and weaknesses in analyzing ophthalmic images.

For anatomical recognition, we employed a comprehensive set of metrics to evaluate the performance of LVLMs:

\textbf{Precision}: Measure the proportion of correctly predicted region types among all predicted regions. A high precision indicates that the model is more likely to be correct when predicting region types. 

\textbf{Recall}: Quantify the proportion of correctly predicted region types among all ground truth regions. A high recall indicates that the model is able to identify a larger fraction of the relevant regions.

\textbf{F1 Score}: The harmonic mean of precision and recall, providing a balanced measure.

\textbf{Hallucination Resistance (HR):} The Hallucination Resistance (HR) metric is a new metric that quantifies a model's ability to avoid hallucinations: 

\begin{equation*} 
\label{eq:hr_metric_cardinality}
\text{HR} = 1 - \frac{|\{r \in \mathcal{P}_i \mid r \notin \mathcal{T}_i\}|}{|\{r \in \mathcal{P}_i \}|},
\end{equation*}

where $\mathcal{P}_i$ represents set of all predicted region IDs for image $i$, and
$\mathcal{T}_i$ indicates set of all ground truth region IDs for image $i$. Higher HR values indicating fewer hallucinations.

These metrics collectively assessed the models' ability to accurately identify and localize anatomical structures in ophthalmic images.

For diagnostic analysis, we focused on the models' performance in glaucoma detection and macular hole staging using the metric of accuracy. It quantified the proportion of correctly classified glaucoma cases and macular hole stages, measuring the models' ability to determine the presence of glaucoma and the severity of macular holes based on visual characteristics.

\begin{figure*}[t!]
    \centering
    \includegraphics[width=1\textwidth]{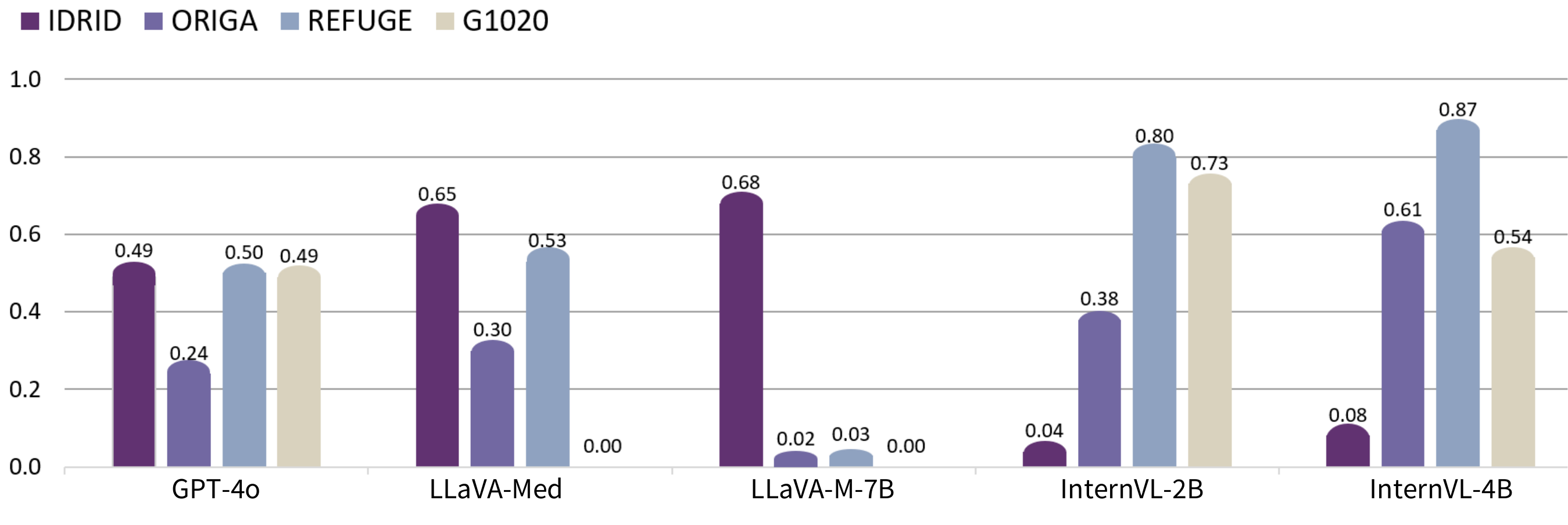}
    \vspace{-20pt}
    \caption{Robustness analysis of LVLMs across different color fundus photograph datasets. The bar chart displays the F1 scores of the five models (GPT-4o, LLaVA-Med, LLaVA-M-7B, InternVL-2B, and InternVL-4B) on four different color fundus photograph datasets: IDRID, ORIGA, REFUGE, and G1020.}
    \label{fig:robust_analysis}
\vspace{-5pt}
\end{figure*}

\vspace{-2pt}
\subsection{Anatomical Recognition}
\vspace{-2pt}

\textbf{Overall Results. }
The results for anatomical recognition in \cref{tab:performance_comparison} revealed that the performance of SoTA LVLMs was far from optimal. The average F1 score across all models was a mere 0.2189, highlighting a significant discrepancy between the models' predictions and the ground truth annotations. The best-performing model, GPT-4o, achieved an F1 score of 0.5748, which, while notably higher than the average, still indicated substantial room for improvement on anatomical recognition. 

\textbf{Nuanced View. }
\cref{fig:radar_chart} provided a more nuanced view of the top five models' F1 performance across different image types. Note that the performance of all models, including GPT-4o, was markedly lower for complex images with a higher number of bounding boxes, such as those in the SS category. This trend suggested that the models struggled to accurately recognize and localize anatomical structures in more intricate and information-dense images. \cref{fig:visual_example} presented visual examples of various responses, including correct and wrong answers, hallucinations, as well as invalid responses. 

\textbf{Robustness Analysis.} The performance distribution of the five best-F1 LVLMs across different CFP datasets, as depicted in \cref{fig:robust_analysis}, revealed variability and inconsistency in model performance. Even for the same image modality, the models exhibited notable differences in performance depending on the specific dataset. For instance, InternVL-4B achieved a high F1 score of 0.87 on REFUGE but experienced a substantial drop in performance on IDRID, with an F1 score of only 0.04. These observations suggested that the models' robustness and generalization ability might be limited when confronted with variations in image quality, acquisition protocols, and patient populations. 

\vspace{-3pt}
\subsection{Diagnosis Analysis}

\textbf{Overall Results. }
The evaluation results for diagnosis analysis in Table~\ref{tab:performance_comparison} demonstrated that the performance was far from perfect. For glaucoma detection, all models achieved accuracies close to random guessing, indicating that they struggled to distinguish between glaucoma and non-glaucoma cases based on the provided ophthalmic images. In the case of macular hole staging, the average accuracy across all models was only 15.31\%, suggesting a significant gap between the models' predictions and the ground truth labels. InternVL-2B, the best-performing model, achieved an accuracy of 30.26\% for macular hole staging. While this number was higher than the average, still fell short of the desired performance for clinical use. 

\textbf{Error Analysis. } 
To better understand the failure modes of LVLMs, we employed the glaucoma diagnosis task as the use case and systematically identified and categorized error types with the assistance of GPT-4o, as shown in Table~\ref{tab:error_analysis}. For each prediction, we provided the predicted result, justification for the prediction, and the ground truth (GT), instructing GPT-4o to categorize errors based on predefined descriptions. Six primary error categories emerged, grounded in established AI evaluation literature: (1) Misclassification \cite{gulshan2016development}; (2) Failure to Abstain \cite{hendrycks2019using}; (3) Inconsistent Reasoning \cite{devlin2019bert}; (4) Hallucination \cite{maynez2020faithfulness}; (5) Assertion without Explanation \cite{rudin2019stop}; (6) Lack of Domain Knowledge \cite{mckinney2020international}. These categories illustrated the models' need for better robustness, domain-specific integration, and uncertainty handling.

\begin{table}
\centering
\resizebox{0.5\textwidth}{!}{
\begin{tabular}{l r r}
\toprule
\textbf{Error Types} & \textbf{Counts} & \textbf{Proportion (\%)} \\
\midrule
Lack of Domain-Specific Knowledge & 3914 & 34.63 \\
Assertion & 2433 & 21.53 \\
Misclassification & 2248 & 19.89 \\
Failure to Abstain & 2035 & 18.01 \\
Hallucination & 439 & 3.88 \\
Inconsistent Reasoning & 232 & 2.05 \\
\bottomrule
\end{tabular}
}
\caption{Statistics of error analysis in glaucoma diagnosis.}
\label{tab:error_analysis}
\vspace{-5pt}
\end{table}

\textbf{E1: Lack of Domain-Specific Knowledge.} The models exhibited a lack of medical knowledge or produced medically inaccurate explanations. For instance, VILA-3B erroneously referred to a macular hole stage 5, claiming that it represents the final stage where the hole has healed and vision is fully restored, a stage that does not exist in the established staging system. This error demonstrated a lack of domain-specific knowledge because the model made incorrect medical assumptions.

\textbf{E2: Assertion.} Models presented assertive predictions without detailed explanations. For example, Yi-6B simply output the glaucoma condition without further explanation, despite being explicitly instructed to provide one. This lack of justification weakens the credibility of the prediction.

\textbf{E3: Misclassification.} Models misidentified conditions or stages, leading to incorrect predictions. For example, LLaVA-Med incorrectly classified non-glaucoma images as glaucoma. 

\textbf{E4: Failure to Abstain.} Some models failed to abstain from making predictions when presented with irrelevant or insufficient data. For example, LLaVA-Med diagnosed glaucoma from an image of a cat, which was clearly not relevant to the medical task. This error type occurred because the model should have recognized that the image was not suitable and refrained from making a prediction.

\textbf{E5: Hallucination.} Models generated details that did not exist in the input image. As illustrated in Figure \ref{fig:visual_example}, LLaVA-M-7B hallucinated an additional region, labeled as region 4, which did not correspond to any ground truth regions. This demonstrates the model's tendency to hallucinate features not present in the actual data.  

\textbf{E6: Inconsistent Reasoning.} Models provided contradictory explanations within their own predictions. For example, InternVL initially predicted a macular hole stage 1 but subsequently  stated  there was no visible macular hole or other abnormalities in the macular region, contradicting its earlier assessment. This inconsistency reflected conflicting reasoning, as the model simultaneously described both a healthy and diseased state.

\textbf{Demographic Subgroup Analysis. } 
We conducted demographic analyses based on age and gender to assess whether the predictive performance of LVLMs is influenced by these factors. Results showed statistically significant differences in accuracy across age and gender subgroups for several models, indicating that certain LVLMs may perform differently across diverse demographic groups. For example, InternVL 2B demonstrated significant variance across both age and gender. See appendix for details.

\vspace{-7pt}
\subsection{Benchmark Justifications}
\vspace{-7pt}

In this section, we justified the design of our benchmark from two perspectives: (1) feasibility and (2) intricacy. These results demonstrated that existing LVLMs faced significant challenges in handling ophthalmic images, even when exposed to medical literature or fine-tuned on relevant datasets.

\textbf{Feasibility}. 
We demonstrated that our proposed benchmark was feasible, as evidenced by the results in Table~\ref{tab:cnn_results}. We showed that supervised neural networks could effectively address both anatomical recognition and diagnosis analysis tasks. For anatomical recognition, we cropped each region and assigned a corresponding label, formulating the task as a multi-class classification problem. For diagnosis analysis, the input consisted of the entire image, with binary labels for glaucoma detection or 4-class labels for MH staging. As shown in Table~\ref{tab:cnn_results}, the classifiers achieved accuracies far above random baselines, justifying the feasibility of our dataset and task formulation. 

\textbf{Intricacy}. 
The poor performance of existing LVLMs on \dataset was not due to the absence of ophthalmic images during pretraining. As shown in Table~\ref{tab:performance_comparison}, LLaVA-Med performed poorly on both anatomical recognition and diagnostic tasks. However, LLaVA-Med had been pre-trained on PubMed, which included extensive medical literature, likely covering ophthalmology topics. More directly, when we presented LLaVA-Med with an optical coherence tomography (OCT) image and a color fundus photograph (CFP), the model correctly identified the image types as ophthalmic images and described the medical conditions, although the diagnoses were incorrect.

Furthermore, fine-tuning did not significantly improve performance. We fine-tuned LLaVA-Med on a combined dataset of OCT and CFP images. To avoid modality dominance, we balanced the number of OCT and CFP images. Following the official fine-tuning protocol \cite{2024_neurips_llava, 2024_neurips_llavamed}, we froze the visual encoder and fine-tuned the MLP adapters and the language model. Despite this, the fine-tuned LLaVA-Med failed to produce meaningful responses as \cref{tab:performance_comparison} showed. For example, it output a series of "opt" for anatomical recognition and empty strings for diagnostic analysis. These findings highlighted that the complexity of ophthalmic images posed inherent challenges that went beyond simple fine-tuning.

\vspace{-2pt}
\section{Limitations}
\vspace{-2pt}

\noindent
\textbf{Modality: }
While \dataset covers five ophthalmic imaging modalities, free-text, and demographic information—making it by far the most comprehensive—it is inevitable that the selected nine datasets may not include all data modalities in ophthalmology. Additionally, longitudinal data is essential for a more thorough assessment of VLLMs in tracking disease progression. These challenges, which are critical, remain open issues in ophthalmology\cite{khan2021global}, and new datasets under development are addressing them\cite{rajesh2023artificial}. We plan to further enhance our data and evaluation pipelines and leverage new datasets to enrich the dataset of \dataset.



\noindent 
\textbf{Environmental Impact}
The evaluation of LVLMs in this study was conducted using NVIDIA H100 GPUs. The energy consumption and carbon footprint associated with training and deploying large-scale AI models have become a growing concern in the research community \cite{2020_green_ai}. It is essential to consider the environmental implications of such practices. The electricity consumed by GPUs during model evaluation contributes to greenhouse gas emissions, depending on the energy mix of the power grid \cite{2020_green_ai}. It is crucial to develop and adopt sustainable practices, as well as promoting the sharing and reuse of pretrained models \cite{2020_green_ai}.

\vspace{-5pt}
\section{Conclusion}
\vspace{-5pt}

In this paper, we introduced \dataset, a comprehensive benchmark designed to evaluate the performance of LVLMs on ophthalmic images, free-text, demographic, and disease biomarker information. Spanning a wide range of image modalities and enriched with annotations for anatomical structures and diagnostic labels, \dataset offers a framework for assessing LVLM capabilities in ophthalmology. Our evaluation of 13 state-of-the-art LVLMs revealed significant shortcomings in understanding ophthalmic images, with models struggling in anatomical recognition, showing inconsistent performance across datasets, and performing close to random in diagnostic tasks such as glaucoma detection and macular hole staging. Fine-tuning on ophthalmic data failed to improve results, highlighting the complexity of these images for LVLMs. In contrast, supervised neural networks trained on the same tasks achieved high accuracy, demonstrating that the challenges are not insurmountable. Our error analysis uncovered six key failure modes of the LVLMs, emphasizing the need for more robust models, better integration of domain-specific knowledge, and improved uncertainty handling. We also showed inconsistencies between subgroups of ages and genders. These discoveries suggest Pressing need for benchmarks in the development and validation of ophthalmology-specific LVLMs. 
\par
\vspace{5pt}
\noindent\textbf{Acknowledgement}. This research is supported by 4R00LM014024, National Library of Medicine, National Institutes of Health.



\bibliography{refs}

\clearpage
\newpage
\appendix

\label{sec:appendix}
\section{Prompts for Benchmarking}
\label{sec:appendix_prompts}

Below are the specific prompts used for the evaluation of large vision-language models (LVLMs) in the tasks of anatomical recognition and diagnosis analysis:

\subsection{Anatomical Recognition Prompts}
The following prompt was used to evaluate the models' ability to recognize and classify anatomical regions in ophthalmic images based on labeled bounding boxes:

\begin{itemize}[left=0pt]
    \setlength\itemsep{0em}
    \item \textbf{Prompt 1:} "This is an image [IMAGE DESCRIPTION] of type [IMAGE TYPE]. Please identify the type of each labeled bounding box in this image. Options can be: [REGION TYPE 1], [REGION TYPE 2], ... Please just follow the format: Region ID: xxx; Type: xxx."
\end{itemize}

\subsection{Diagnosis Analysis Prompts}
The following prompts were used to assess the models' diagnostic reasoning for conditions such as glaucoma and macular hole staging:

\begin{itemize}[left=0pt]
    \setlength\itemsep{0em}
    \item \textbf{Prompt 1:} "This is <IMAGE TYPE>. Based on the image, please tell me whether this image contains glaucoma, then give justifications. Follow the format: GLAUCOMA / NON-GLAUCOMA; Explanation: <JUSTIFICATION>."
    
    \item \textbf{Prompt 2:} "This is <IMAGE TYPE>. Based on the image, please tell me the stage of the macular hole, then give justifications. Follow the format: Stage: <AN INTEGER>; Explanation: <JUSTIFICATIONS>."
\end{itemize}

\section{Intrinsic Difficulties of Ophthalmology}

\begin{figure*}
    \centering
    \includegraphics[width=\linewidth]{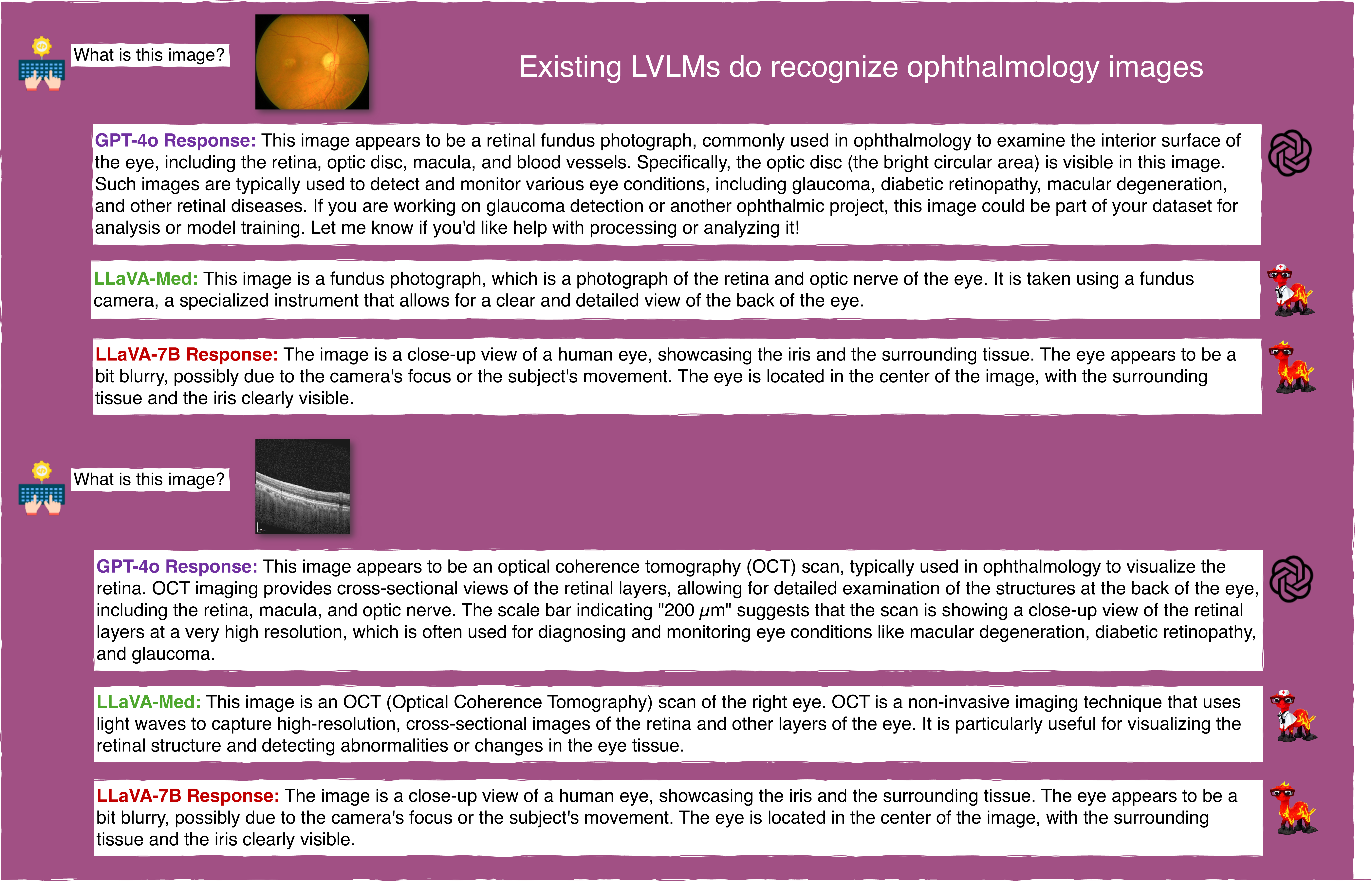}
    \caption{Responses from GPT-4o and LLaVA-7B when presented with a retinal fundus photograph and asked, ``What is this image?'' }
    \label{fig:ophthalmology_recognition}
\end{figure*}

While the low performance of LVLMs on the \dataset benchmark might suggest a lack of exposure to ophthalmology images during pretraining, our investigation reveals that these models can indeed recognize such images. As demonstrated in \cref{fig:ophthalmology_recognition}, when presented with a retinal fundus photograph and asked to identify the image, both GPT-4o and LLaVA-7B correctly recognize it as an ophthalmology-related image. GPT-4o specifically identifies the image as a ``retinal fundus photograph, commonly used in ophthalmology to examine the interior surface of the eye,'' highlighting its potential applications in detecting and monitoring various eye conditions. Similarly, LLaVA-7B recognizes the image as a close-up view of a human eye, albeit focusing more on superficial features such as the iris and surrounding tissue. These findings suggest that LVLMs have been exposed to ophthalmology data during training and can identify such images when encountered. However, the performance of LVLMs on more complex tasks, such as glaucoma classification, remains a challenge, as evidenced by the results on the \dataset benchmark. This discrepancy between image recognition and task-specific performance underscores the inherent challenges posed by ophthalmology data for LVLMs.

Regarding the first point, we find that LVLMs like GPT-4o and LLaVA-7B can correctly identify color fundus photographs (CFPs) from \dataset, indicating that these models have encountered ophthalmology data during training. This suggests that the low performance is not due to a complete lack of exposure to the domain.

As for the second point, we fine-tuned LLaVA-7B on the glaucoma detection task using the training set of \dataset. Notably, even on the same training data, the fine-tuned model still achieves only 50\% accuracy, always predicting "non-glaucoma" regardless of the input. This indicates that simple fine-tuning is insufficient to overcome the random accuracy in glaucoma detection, underscoring the inherent challenges of ophthalmology data for LVLMs.

Additionally, some may question the suitability of the ophthalmology images in \dataset for glaucoma classification. However, this concern is unfounded. As demonstrated in \cite{2023_origa_cls}, a deep learning model achieves an accuracy of 93.86\% on the ORIGA dataset, which is the same dataset used for evaluating glaucoma detection in our benchmark. This high performance indicates that the images in ORIGA contain sufficient discriminative information for distinguishing between glaucoma and non-glaucoma cases. Consequently, the poor performance of LVLMs on this task cannot be attributed to the quality or suitability of the dataset itself, but rather highlights the inherent challenges that ophthalmology data poses for these models.

\section{Experimental Setup}
We developed a general framework based on PyTorch, providing a unified interface for performing inference across various vision-language models (VLMs). This framework ensures consistent evaluation and smooth integration with different models.

For each VLM, we used the same computing infrastructure—specifically, two RTX 6000 GPUs—to perform the inference. We evaluated the models using nine different ophthalmology datasets, with identical prompts and inputs provided to each VLM. Moreover, we applied the default hyperparameters for each model during the evaluation. This approach allowed us to fairly compare the performance of the different models.


\section{Computational Resource}
The computing infrastructure includes 11 GPU nodes, each equipped with 2x AMD EPYC 7742 processors (128 cores), 1TB of RAM, and 8 Quadro RTX 6000 GPUs per node. Additionally, there are 7 GPU nodes with 2x Intel Icelake Xeon Platinum 8358 processors.

For vision-language model inference tasks on various ophthalmology datasets, the runtime typically ranges from two to four hours, depending on the specific dataset.

\section{Use Of AI Assistants}
We used AI tools to assist with coding tasks, such as debugging and optimizing code during the development phase. Additionally, we leveraged AI to help polish the manuscript, addressing grammar issues and ensuring clarity and coherence in our presentation. However, all critical decisions such as the research design, methodology, and conclusions were made independently by the authors.

\section{Hyperparameters}

This section outlines the essential hyperparameters that were chosen for the large vision-language models (LVLMs) in our experiments, listed in Table~\ref{tab:hyperparameters}. 

\begin{enumerate}

\item \textbf{Image Resolution}: The image resolution defines the size of the visual input processed by each LVLM. Higher resolutions capture finer details. Table~\ref{tab:hyperparameters} summarizes the exact values of these hyperparameters.

\item \textbf{Top-p Sampling}: Top-p, also known as nucleus sampling, is a hyperparameter that influences the randomness of a language model's output. It defines a probability threshold and selects the smallest set of tokens whose cumulative probability exceeds this threshold. The model then samples randomly from this subset to generate the output. This approach allows for more diverse and creative results compared to methods that randomly sample from the entire vocabulary.

\item \textbf{Temperature}: The temperature hyperparameter influences the randomness of the model's output by scaling logits before applying softmax. Higher temperatures (e.g., >1) encourage more diverse outputs by flattening the probability distribution, making it suitable for creative tasks. Lower temperatures (e.g., <1) concentrate the distribution, resulting in more focused outputs, which is critical in medical domains to ensure reliable, deterministic responses. Temperature is disabled when setting to be 0. 

\item \textbf{Beams Number}: Beam search is a decoding strategy that retains multiple candidate sequences at each generation step. A higher number of beams (e.g., 5 or 10) explores more possibilities, potentially yielding better results at the cost of increased computation. A lower beams number (e.g., 1) favors efficiency and speed but risks missing better sequences, which may be a concern in domains requiring high-quality outputs.

\item \textbf{Number of Parameters}: The number of parameters refers to the total count of learnable weights in a model, directly influencing its capacity and performance. Larger models tend to perform better due to increased capacity, but at the cost of higher memory usage and slower inference times.

\item \textbf{Max New Tokens}: This hyperparameter limits the number of tokens generated by the model during inference.

\end{enumerate}

\begin{table*}[h!]
\centering
\resizebox{0.8\textwidth}{!}{
\begin{tabular}{lccccrc}
\toprule
\textbf{Model} & \textbf{Image} & \textbf{Top-p} & \textbf{Temperature} & \textbf{Beams} & \textbf{Number of} & \textbf{Max New} \\
               & \textbf{Resolution} & \textbf{Sampling} &                    & \textbf{Number} & \textbf{Parameters} & \textbf{Tokens} \\
\midrule
GPT-4o         & 512 x 512   & -     & -    & -     & -    & 512  \\
LLaVA-Med      & 336 x 336   & 1.0   & 0    & 1     & 7.56 B   & 512  \\
LLaVA-1.5-7B   & 336 x 336   & 1.0   & 0    & 1     & 7.06 B & 512  \\
LLaVA-M-7B     & 336 x 336   & 1.0   & 0    & 1     & 7.56 B & 512  \\
LLaVA-V-7B     & 336 x 336   & 1.0   & 0    & 1     & 7.06 B   & 512  \\
LLaVA-13B      & 336 x 336   & 1.0   & 0    & 1     & 13.35 B  & 512  \\
Yi 6B          & 336 x 336   & 1.0   & 0    & 1     & 6.71 B   & 512  \\
InternVL 2B    & 448 x 448   & -     & 0    & 1     & 2.20 B   & 512  \\
InternVL 4B    & 448 x 448   & -     & 0    & 1     & 4.14 B   & 512  \\
QWen           & 448 x 448   & -     & 0    & -     & 9.65 B & 512  \\
VILA 3B        & 384 x 384   & 1.0   & 0    & 1     & 3.14 B   & 512  \\
VILA 3B-S2     & 768 x 768   & 1.0   & 0    & 1     & 3.16 B   & 512  \\
VILA 8B        & 384 x 384   & 1.0   & 0    & 1     & 8.49 B   & 512  \\
\bottomrule
\end{tabular}
}
\caption{Hyperparameters for the large vision-language models evaluated in this work. The ``Image Resolution'' column shows the spatial dimensions of the input images processed by each model. The ``Top-p Sampling'' column indicates the value of the top-p sampling hyperparameter used during text generation. The ``Temperature'' column specifies the value of the temperature hyperparameter used in the softmax function during text generation. It is disabled given value 0. The \``Beams Number'' column indicates the number of beams used in the beam search decoding algorithm. The ``Number of Parameters'' column reports the model size in billions (B) of parameters. The ``Max New Tokens'' column indicates the maximum number of new tokens the model can generate during inference. Symbol ``-'' denoting cases where it is not applicable or not reported.}
\label{tab:hyperparameters}
\end{table*}
\section{Response Examples}
\label{sec:response_examples}

\begin{tcolorbox}[colframe=inputcolor, colback=inputcolor!20, width=\columnwidth, title=\textcolor{black}{\textbf{Input:}}, breakable]
\texttt{This is an ophthalmology OCT image. Based on the image, please tell me the stage of macular hole decision. Then, give a detailed justification and explanation for your answer. Follow the format: \{Stage: <AN INTEGER>; Justification: <EXPLANATION>\}.}
\end{tcolorbox}

\begin{tcolorbox}[colframe=gtcolor, colback=gtcolor!20, width=\columnwidth, title=\textcolor{black}{\textbf{Ground Truth (GT):}}, breakable]
\texttt{Stage: 4}
\end{tcolorbox}

\begin{tcolorbox}[colframe=outputcolor, colback=outputcolor!20, width=\columnwidth, title=\textcolor{black}{\textbf{GPT-4o Response:}}, breakable]
\texttt{Stage: 4; Justification: The OCT image displays a full-thickness macular hole, where there is a complete defect involving all the retinal layers from the inner surface to the retinal pigment epithelium. The presence of this full-thickness defect, along with the evident central gap extending through the macula and alterations in the surrounding retina, clearly indicates a Stage 4 macular hole. Additionally, the involvement of the posterior vitreous detachment supports this stage classification.}
\end{tcolorbox}

\begin{tcolorbox}[colframe=outputcolor, colback=outputcolor!20, width=\columnwidth, title=\textcolor{black}{\textbf{LLaVA Med Response:}}, breakable]
\texttt{Stage: 3; Justification: The image shows a macular hole with a diameter of 300 micrometers. This is considered stage 3, which is a more advanced stage of macular hole development. The macular hole is larger and may be associated with more severe vision loss.}
\end{tcolorbox}

\begin{tcolorbox}[colframe=outputcolor, colback=outputcolor!20, width=\columnwidth, title=\textcolor{black}{\textbf{LLaVA-1.5-7B Response:}}, breakable]
\texttt{Stage: 2; Justification: The image shows a macular hole with a well-defined, clean, and clear appearance. The hole is surrounded by a well-organized retinal structure, including the fovea and the outer retinal layers. The presence of a clear and well-defined macular hole indicates that it is in an early stage of development, which is typically characterized by a clean and organized appearance.}
\end{tcolorbox}

\begin{tcolorbox}[colframe=outputcolor, colback=outputcolor!20, width=\columnwidth, title=\textcolor{black}{\textbf{LLaVA-M-7B Response:}}, breakable]
\texttt{The image you've provided appears to be an Ophthalmology OCT (Optical Coherence Tomography) scan, which is a non-invasive imaging technique used to visualize the layers of the retina...}
\end{tcolorbox}

\begin{tcolorbox}[colframe=outputcolor, colback=outputcolor!20, width=\columnwidth, title=\textcolor{black}{\textbf{LLaVA-V-7B Response:}}, breakable]
\texttt{Stage: 200 \textmu m; Justification: The macular hole is not clearly visible in this image. The hole is typically smaller than 200 \textmu m, which is the scale indicated by the measurement on the left side of the image.}
\end{tcolorbox}

\begin{tcolorbox}[colframe=outputcolor, colback=outputcolor!20, width=\columnwidth, title=\textcolor{black}{\textbf{LLaVA-13B Response:}}, breakable]
\texttt{Stage: 2; Justification: The image shows a macular hole in the retina, which is a stage 2 macular hole. In stage 2, the hole is small and has not yet begun to contract.}
\end{tcolorbox}

\begin{tcolorbox}[colframe=outputcolor, colback=outputcolor!20, width=\columnwidth, title=\textcolor{black}{\textbf{Yi-6B Response:}}, breakable]
\texttt{Stage: 4}
\end{tcolorbox}

\begin{tcolorbox}[colframe=outputcolor, colback=outputcolor!20, width=\columnwidth, title=\textcolor{black}{\textbf{InternVL-2B Response:}}, breakable]
\texttt{Stage: 2; Justification: The image shows a macular hole, which is a small hole in the retina that can cause vision loss.}
\end{tcolorbox}

\begin{tcolorbox}[colframe=outputcolor, colback=outputcolor!20, width=\columnwidth, title=\textcolor{black}{\textbf{InternVL-4B Response:}}, breakable]
\texttt{Stage: 1; Justification: The image shows a clear view of the retinal layers with a distinct, well-defined macular hole.}
\end{tcolorbox}

\begin{tcolorbox}[colframe=outputcolor, colback=outputcolor!20, width=\columnwidth, title=\textcolor{black}{\textbf{QWen Response:}}, breakable]
\texttt{Stage: 3; Justification: The image shows a large, full-thickness macular hole with posterior vitreous detachment and subretinal fluid.}
\end{tcolorbox}

\begin{tcolorbox}[colframe=outputcolor, colback=outputcolor!20, width=\columnwidth, title=\textcolor{black}{\textbf{VILA-8B Response:}}, breakable]
\texttt{[0.00, 0.00, 1.00, 1.00]}
\end{tcolorbox}

\section{Supervised Training Settings}

To evaluate the feasibility of our proposed benchmark, we implemented neural network classifiers for anatomical recognition and diagnosis analysis. For both tasks, we used 80\%, 15\%, and 5\% data for training, validation, and test. For anatomical recognition, we employed a CNN visual encoder whose architecture is like below: 

\begin{figure*}[t]
\centering
\begin{minipage}{0.9\textwidth}
\begin{lstlisting}[language=Python, frame=single, basicstyle=\ttfamily\footnotesize]
class RegionClassifier(nn.Module):
    def __init__(self, num_classes):
        self.conv1 = nn.Conv2d(3, 32, kernel_size=3, stride=1, padding=1)
        self.pool = nn.MaxPool2d(2, 2)
        self.conv2 = nn.Conv2d(32, 64, kernel_size=3, stride=1, padding=1)
        self.fc1 = nn.Linear(64 * 32 * 32, 128)
        self.fc2 = nn.Linear(128, num_classes)
        self.relu = nn.ReLU()

    def forward(self, x):
        x = self.pool(self.relu(self.conv1(x)))
        x = self.pool(self.relu(self.conv2(x)))
        x = x.view(-1, 64 * 32 * 32)
        x = self.relu(self.fc1(x))
        x = self.fc2(x)
        return x
\end{lstlisting}
\end{minipage}
\caption{The RegionClassifier model implementation in PyTorch.}
\label{fig:region_classifier}
\end{figure*}

\noindent The CNN was trained with the following settings:
\begin{itemize}
    \item Image resolution: 128 $\times$ 128
    \item Batch size: 512
    \item Learning rate: 0.001
    \item Epochs: 20
\end{itemize}

For diagnostic analysis, we fine-tuned RETFound as the visual encoder. RETFound is a foundation model for retinal images, built on a large Vision Transformer (ViT) architecture with 24 Transformer blocks and an embedding vector size of 1,024 \cite{2023_nature_eye_foundation}. The RETFound model offers two variations designed for different image types: CFP 
and OCT.
For macular hole (MH) stage classification, we employed the OCT variation, while the CFP model was used for glaucoma classification (according to the dataset's image type).
For both tasks, we fine-tuned RETFound using the default parameter settings:
\begin{itemize}
    \item Image resolution: 224 $\times$ 224
    \item Batch size: 16
    \item Base learning rate:: 5e-3
    \item Epochs: 50
    \item Layer decay: 0.65 
    \item Weight decay: 0.05 
    
\end{itemize}

The model's performance on anatomical recognition and diagnosis analysis tasks served as a baseline for the complexity of our dataset, and is compared with the performance of LVLMs in subsequent sections.

\begin{figure}[tp]
    \centering
    \includegraphics[width=0.99\linewidth]{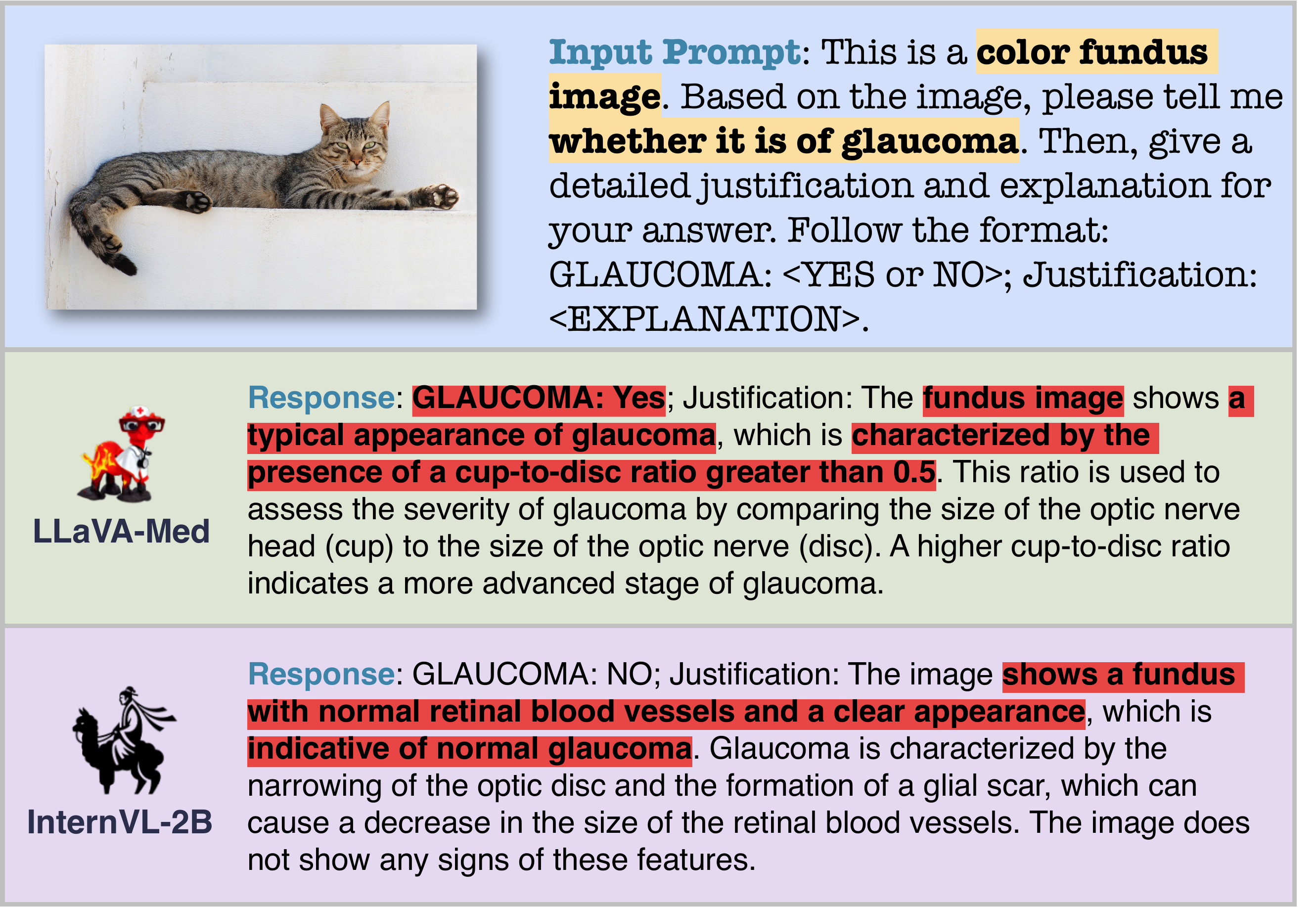}
    \caption{\textbf{Failure to abstain from making predictions on out-of-domain images.} The figure illustrates the responses generated by LLaVA-Med and InternVL-2B when presented with a non-ophthalmology image of a cat and queried about the presence of glaucoma. Despite the image being outside the domain of fundus photography, LLaVA-Med incorrectly classifies the cat image as showing signs of glaucoma, citing a typical cup-to-disc ratio greater than 0.5. Similarly, InternVL-2B misinterprets the cat image as a fundus image but concludes that there are no signs of glaucoma based on the visible features. }
    \label{fig:blind}
\end{figure}
\begin{figure}[tp]
    \centering
    \includegraphics[width=0.99\linewidth]{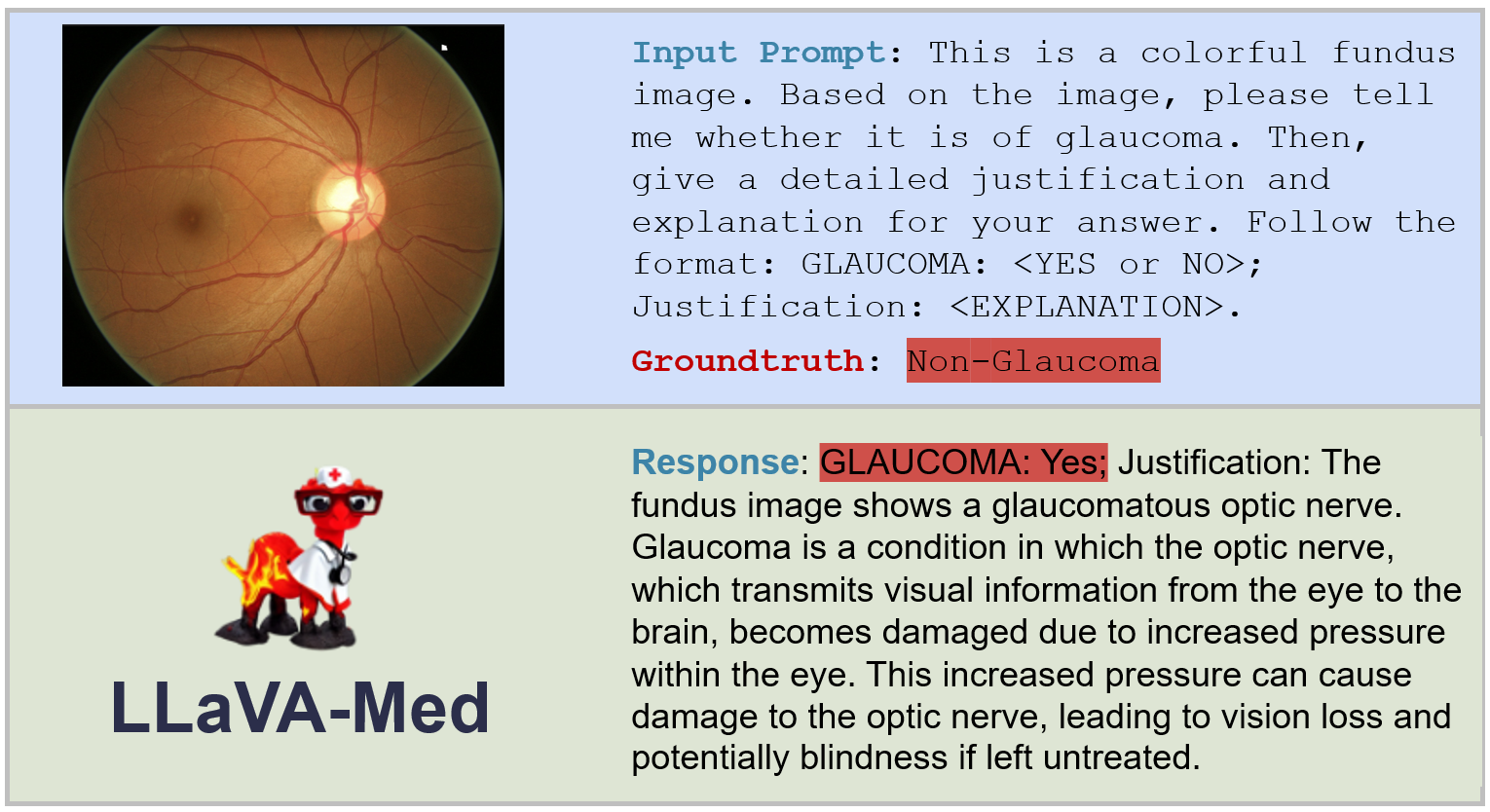}
    \caption{LLaVA-Med incorrectly predicted the image as glaucoma, providing a detailed but inaccurate explanation.}
    \label{fig:misclassification}
\end{figure}
\begin{figure}[tp]
    \centering
    \includegraphics[width=0.99\linewidth]{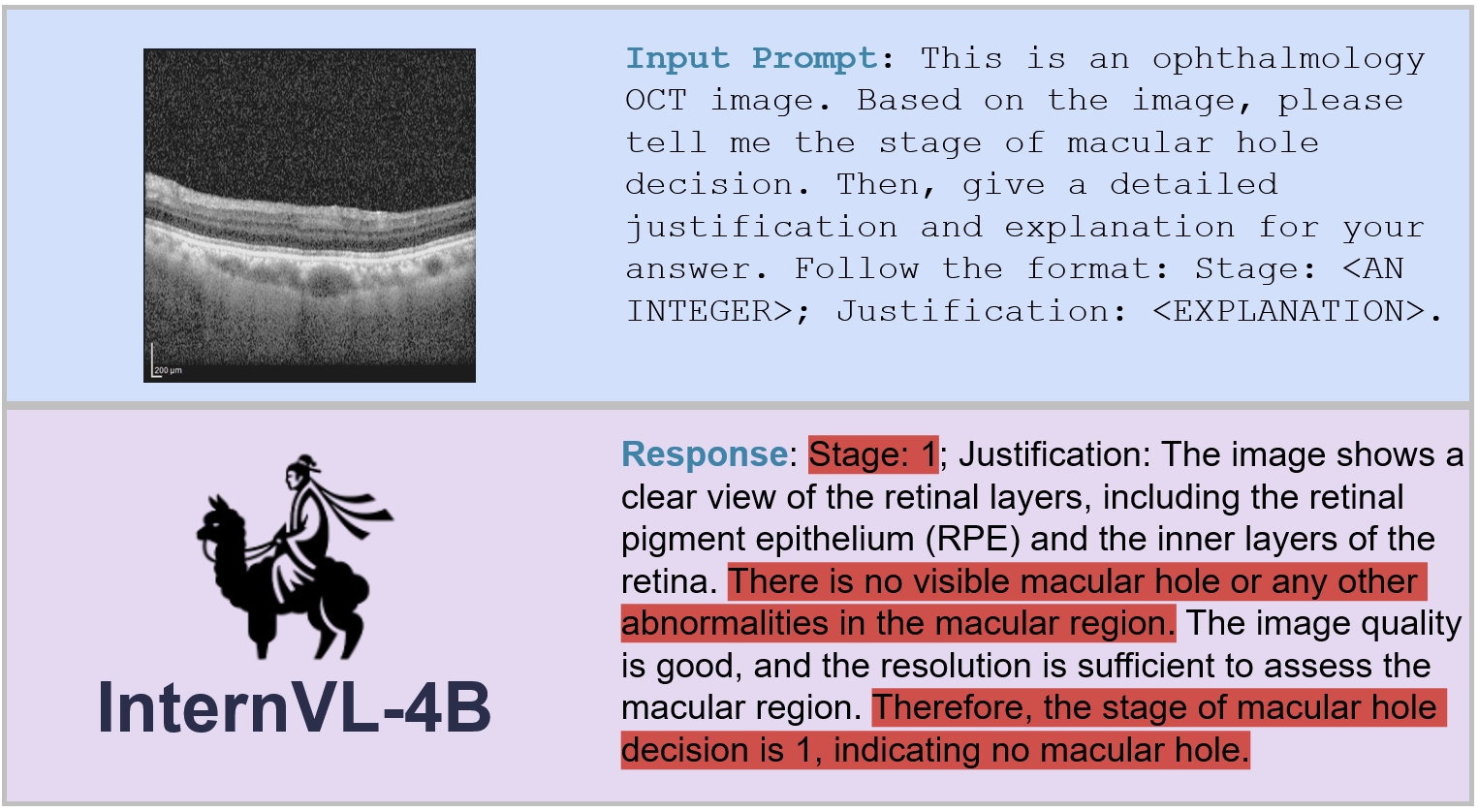}
    \caption{InternVL-4B presented contradictory explanations within its prediction.}
    \label{fig:inconsistency}
\end{figure}
\begin{figure}[tp]
    \centering
    \includegraphics[width=0.99\linewidth]{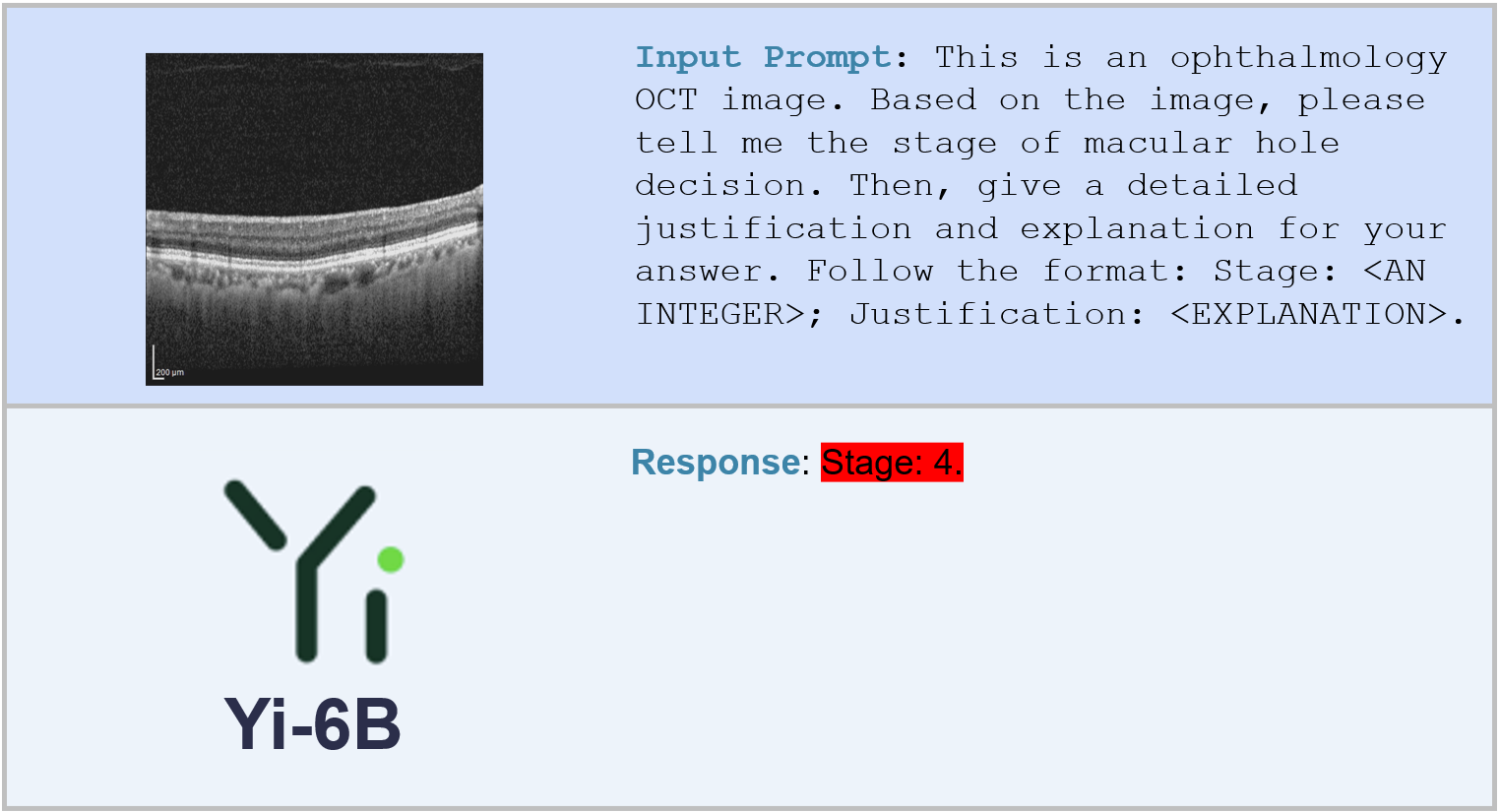}
    \caption{Yi-6B provided assertive predictions without offering detailed explanations.}
    \label{fig:assertion}
\end{figure}
\begin{figure}[tp]
    \centering
    \includegraphics[width=0.99\linewidth]{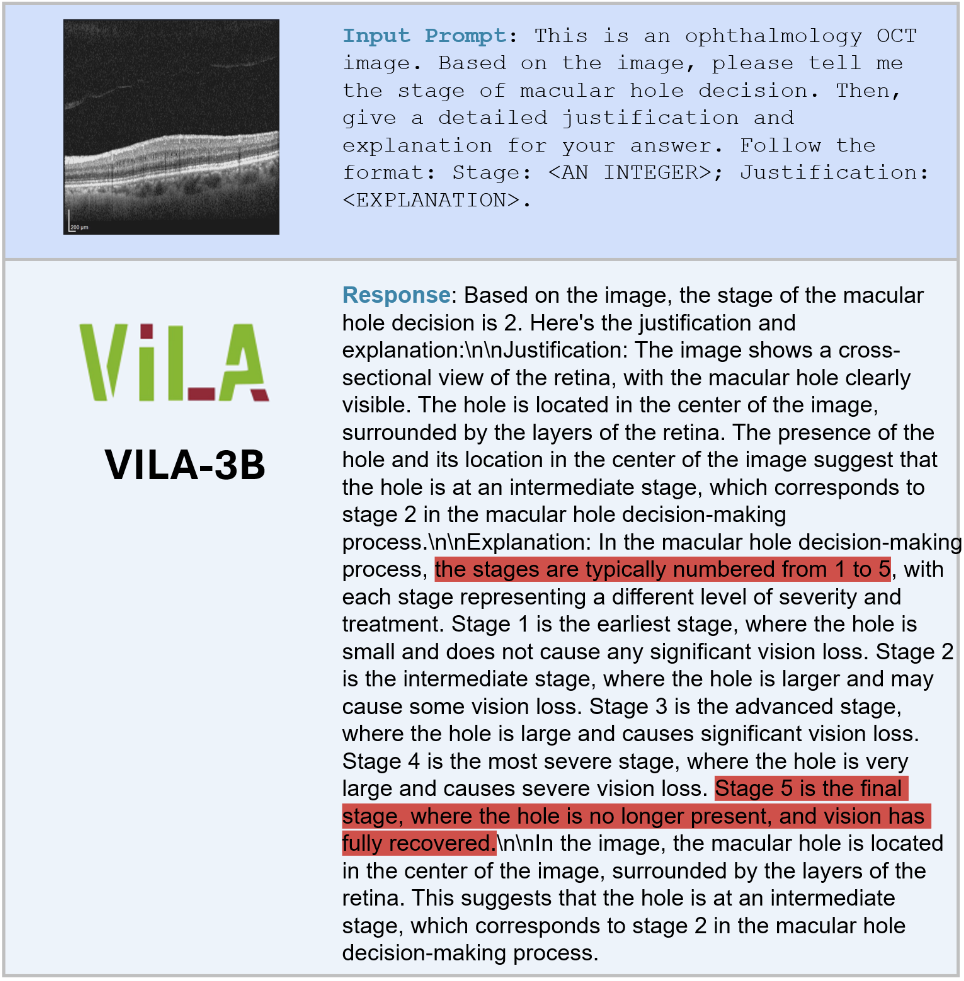}
    \caption{LVLMs display a lack of medical knowledge and produce medically inaccurate explanations. The figure illustrates the responses generated by VILA-3B}
    \label{fig:domain_knowledge}
\end{figure}

\section{Detailed Error Analysis}

\subsection{Misclassification}
As shown in \cref{fig:misclassification}, LLaVA-Med incorrectly predicted the image as glaucoma, providing a detailed but inaccurate explanation. Experimental results in Table \ref{tab:performance_comparison} indicate that all LVLMs achieve suboptimal accuracy in both glaucoma and macular hole stage classification tasks. Additionally, models in the LLaVA series, such as LLaVA-Med, LLaVA-13B, and LLaVA-M-7B, consistently generate the same response, predicting glaucoma disease across different input images. These findings suggest that current LVLMs still lack the capability to accurately interpret images and make reliable predictions.

\subsection{Failure to Abstain}
Some models failed to recognize when the input image was irrelevant to the medical task at hand and continued to provide diagnostic predictions regardless of the mismatch. As \cref{fig:blind} showed, LLaVA-Med and InternVL-2B misdiagnosed the image of a cat as showing signs of glaucoma. In the provided scenario, the prompt explicitly asked whether the image depicted a case of glaucoma, which typically involves human ophthalmic images, such as retinal scans or fundus photographs. However, the input image was clearly a photograph of a cat, an object entirely outside the scope of the medical context. The inability of models like LLaVA-Med to abstain from making predictions on out-of-domain inputs points to their limitations in robustly handling uncertainty or recognizing when data does not meet the conditions required for valid predictions. It underscores the need for integrating mechanisms into LVLMs that can detect when the input data is irrelevant, ensuring that models avoid generating misleading or incorrect medical diagnoses.

\subsection{Inconsistent Reasoning}
As observed in \ref{fig:inconsistency}, InternVL-4B initially predicted the macular hole stage of the input image as stage 1 but then stated in its explanation that there was no visible macular hole or other abnormalities in the macular region, ultimately concluding with no macular hole. Consistent reasoning is proven to be crucial for providing accurate answers, yet this explanation contradicts the initial prediction, highlighting the intrinsic issue of inconsistent model reasoning.

\subsection{Assertion}
Some LVLMs, such as Yi-6B, tend to provide direct answers without offering detailed explanations. As shown in \ref{fig:assertion}, Yi-6B simply predicted the presence of glaucoma without providing any justification, despite the prompt explicitly requesting one. In practical clinical scenarios, reasoning steps are critically important. AI models should not only assist clinicians by streamlining their workflow but also ensure decision-making transparency. This allows clinicians to validate both the final prediction and the intermediate reasoning steps, helping to identify potential issues and fostering trust in AI-assisted clinical systems.

\subsection{Lack of Domain-Specific Knowledge}
Some Large Vision-Language Models (LVLMs) demonstrate inadequate medical knowledge and produce medically inaccurate explanations. As illustrated in Figure \ref{fig:domain_knowledge}, VILA-3B inaccurately performs an initial analysis by misclassifying the stages of macular holes, erroneously numbering them from 1 to 5. Specifically, it incorrectly identifies Stage 5 as the final stage, where the macular hole is presumed to be resolved and vision fully restored. In reality, the established staging system for macular holes ranges from Stage 1 to Stage 4, without Stage 5. These inaccuracies highlight the persistent issue of LVLMs lacking domain-specific expertise.

\begin{figure}[htp!]
    \centering
    \includegraphics[width=0.8\linewidth]{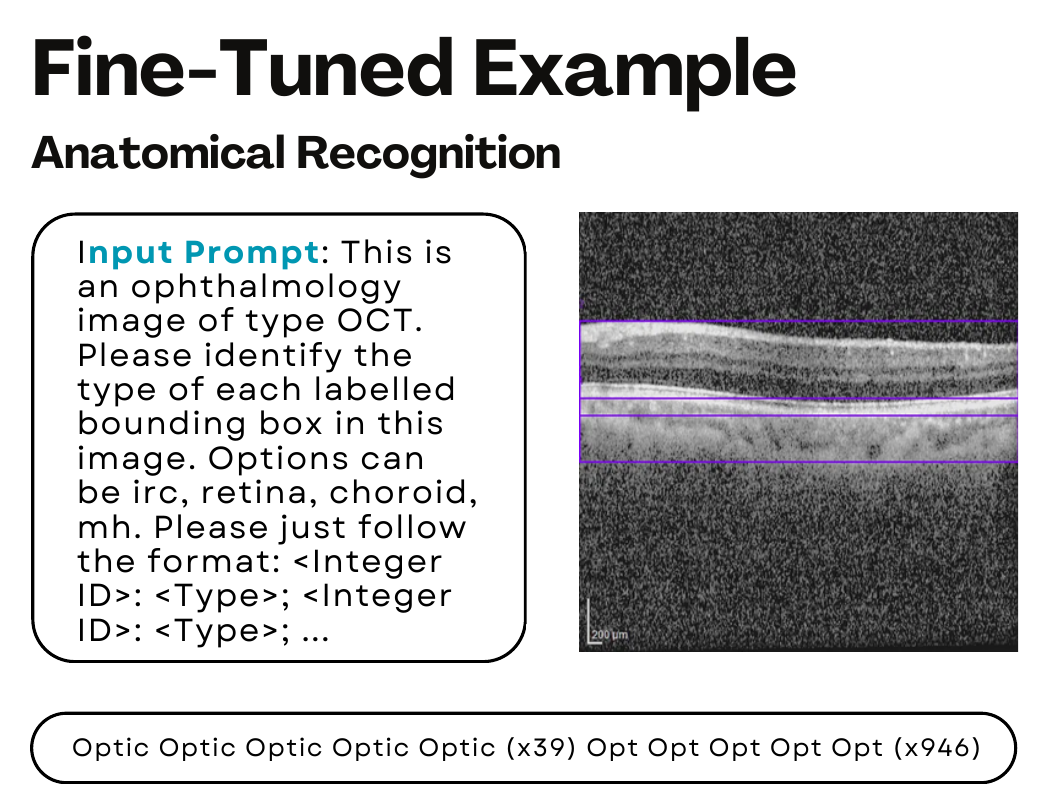}
    \caption{Fine-tuned example response for anatomical recognition. The model was tasked with identifying the type of each labeled region in an OCT image. The fine-tuned LVLM failed to produce meaningful responses, outputting repetitive text such as "Optic Optic Optic Optic Optic (x39)" followed by "Opt Opt Opt Opt Opt (x946)." }
    \label{fig:finetuned_example_anatomical_recognition}
\end{figure}

\section{Details of Fine-tuning LVLM}

We provide the fine-tuning configuration used for training the large vision-language model (LVLM) with the script in the following script. During fine-tuning for anatomical recognition, the model failed to produce coherent outputs, generating the repetitive response: ``Optic Optic Optic Optic Optic (x39) Opt Opt Opt Opt Opt (x946)''. For diagnostic tasks, such as glaucoma and macular hole staging, the model returned no responses. 

\begin{figure*}[ht]
\begin{verbatim}
deepspeed llava/train/train_mem.py \
    --model_name_or_path llava-med-7b-delta \
    --data_path <DATA_PATH> \
    --vision_tower openai/clip-vit-large-patch14 \
    --mm_vision_select_layer -2 \
    --mm_use_im_start_end True \
    --bf16 True \
    --output_dir <OUTPUT_PATH> \
    --num_train_epochs 3 \
    --per_device_train_batch_size 1 \
    --per_device_eval_batch_size 4 \
    --gradient_accumulation_steps 8 \
    --evaluation_strategy "no" \
    --save_strategy "steps" \
    --save_steps 5000 \
    --save_total_limit 3 \
    --learning_rate 2e-5 \
    --weight_decay 0. \
    --warmup_ratio 0.03 \
    --lr_scheduler_type "cosine" \
    --logging_steps 1 \
    --tf32 True \
    --fsdp "full_shard auto_wrap" \
    --fsdp_transformer_layer_cls_to_wrap 'LlamaDecoderLayer' \
    --model_max_length 2048 \
    --gradient_checkpointing True \
    --lazy_preprocess True \
    --report_to wandb
\end{verbatim}
\end{figure*}

\section{Demographic analysis}

\begin{table}
\centering
\resizebox{0.2\textwidth}{!}{
\begin{tabular}{c|c}
\toprule
\textbf{Model} & \textbf{P-value} \\ 
\midrule 
GPT4o & 0.3395 \\ \hline
LLaVA-Med & 5.1625e-15 \\ \hline
LLaVA-1.5-7B & 9.3736e-06 \\ \hline
LLaVA-M-7B & 0.1766 \\ \hline
LLaVA-V-7B & N/A \\ \hline
LLaVA-13B & N/A \\ \hline
Yi 6B & 0.1022 \\ \hline
InternVL 2B & 0.0472 \\ \hline
InternVL 4B & 0.0754 \\ \hline
Qwen & 9.0541e-16 \\ \hline
VILA 3B & 0.5818 \\ \hline
VILA 3B-S2 & 0.7814 \\ \hline
VILA 8B & 0.9269 \\ 
\bottomrule 
\end{tabular}
}
\vspace{-1mm}
\captionsetup{font=small}
\caption{P-values from the age group analysis of various LVLMs.}
\label{tab:age_pvalue}
\vspace{-3mm}
\end{table}

\begin{table}
\centering
\resizebox{0.25\textwidth}{!}{
\begin{tabular}{c|c}
\toprule
\textbf{Model} & \textbf{P-value} \\ 
\midrule 
GPT4o & 0.3734 \\ \hline
LLaVA-Med & 0.4411 \\ \hline
LLaVA-1.5-7B & 8.5040e-07 \\ \hline
LLaVA-M-7B & 0.1937 \\ \hline
LLaVA-V-7B & N/A \\ \hline
LLaVA-13B & N/A \\ \hline
Yi 6B & 0.1681 \\ \hline
InternVL 2B & 6.9497e-24 \\ \hline
InternVL 4B & 0.0082 \\ \hline
Qwen & 0.8096 \\ \hline
VILA 3B & 0.9345 \\ \hline
VILA 3B-S2 & 0.1090 \\ \hline
VILA 8B & 0.2258 \\ 
\bottomrule 
\end{tabular}
}
\vspace{-1mm}
\captionsetup{font=small}
\caption{P-values from the gender group analysis of various LVLMs.}
\label{tab:gender_pvalue}
\end{table}


Tables \ref{tab:age_pvalue} and \ref{tab:gender_pvalue} provide detailed p-value statistics from the age and gender subgroup analyses of various large vision-language models (LVLMs). Specifically, we evaluated the models across different age groups (18-40, 40-60, 60+) and gender categories (male, female). This two-tiered evaluation approach allowed us to examine how demographic factors such as age and gender influence the predictive accuracy of the models. Statistically significant differences were observed for several models, including the following examples:

\begin{itemize}
    \item \textbf{InternVL 2B}: $p = 0.0472$ for age, and $p = 6.94 \times 10^{-24}$ for gender.
    \item \textbf{LLaVA-1.5-7B}: $p = 9.3736 \times 10^{-6}$ for age, and $p = 8.5040 \times 10^{-7}$ for gender.
    \item \textbf{LLaVA-Med} and \textbf{Qwen}: significant $p$-values in the age subgroup analysis ($p = 5.16 \times 10^{-15}$ and $p = 9.05 \times 10^{-16}$, respectively).
    \item \textbf{InternVL 4B}: significant $p$-value in the gender subgroup analysis ($p = 0.0082$).
\end{itemize}

\section{Detailed benchmarked LVLMs}
We benchmarked 13 LVLMs on the \dataset benchmark. These models included:

\noindent
\textbf{GPT-4o}: A proprietary model developed by OpenAI \cite{2023_openai_gpt4}.

\noindent
\textbf{LLaVA Variations}: LLaVA \cite{2024_neurips_llava} leveraged a pre-trained vision encoder and a large language model to achieve state-of-the-art performance on various vision-language tasks. We evaluated several variations based on different language models and sizes, including LLaVA-7B, LLaVA-M-7B (M for Mistral), LLaVA-V-7B (V for Vicuna), LLaVA-13B, and LLaVA-Med \cite{2024_neurips_llavamed}, which was fine-tuned on a large-scale medical image-text dataset.

\noindent
\textbf{Yi-6B}: A vision-language model supporting both Chinese and English \cite{2024_yi}.

\noindent
\textbf{InternVL Variations}: InternVL \cite{2024_internvl_cvpr} aligned a scaled-up vision foundation model with a LLM using web-scale image-text data. We evaluated two variants with 2B and 4B parameters. 

\noindent
\textbf{QWen}: A vision-language model that used a query-based approach to align visual and textual representations \cite{bai2023qwen}.

\noindent
\textbf{VILA Variations}: VILA explored different pre-training strategies for LVLMs. We evaluated three variants: VILA-3B, VILA-3B-S2, and VILA-8B, with parameter counts ranging from 3 billion to 8 billion \cite{2023_vila}.

\end{document}